\documentclass[10pt,journal,compsoc]{IEEEtran}
\usepackage{multirow}
\usepackage{booktabs} 
\usepackage[font=large,labelfont=bf]{caption}
\usepackage{subfigure}
\usepackage{tabularx}
\usepackage{multirow}
\usepackage{array}
\usepackage{color}
\usepackage{amsfonts}
\usepackage{amsmath}
\usepackage{enumitem}
\usepackage{graphicx}
\usepackage{url}
\usepackage{dsfont}
\usepackage{enumerate}
\usepackage{times}
\usepackage{balance}
\usepackage{algorithm}
\usepackage{algorithmic}
\usepackage{hyperref}

\newlist{Properties}{enumerate}{2}
\setlist[Properties]{label=Property \arabic*.,itemindent=*}

\ifCLASSOPTIONcompsoc
  \usepackage[nocompress]{cite}
\else
  \usepackage{cite}
\fi

\begin{document}
\title{Spectral Adversarial Training for Robust \\Graph Neural Network}
\author{Jintang Li, Jiaying Peng, Liang Chen*\thanks{*Corresponding author. E-mail: chenliang6@mail.sysu.edu.cn}, Zibin Zheng, Tingting Liang, Qing Ling}
\IEEEtitleabstractindextext{%
    \begin{abstract}
        Recent studies demonstrate that Graph Neural Networks (GNNs) are vulnerable to slight but adversarially designed perturbations, known as \emph{adversarial examples}. To address this issue, robust training methods against adversarial examples have received considerable attention in the literature. \emph{Adversarial Training (AT)} is a successful approach to learning a robust model using adversarially perturbed training samples. Existing AT methods on GNNs typically construct adversarial perturbations in terms of graph structures or node features. However, they are less effective and fraught with challenges on graph data due to the discreteness of graph structure and the relationships between connected examples.
        In this work, we seek to address these challenges and propose \textbf{S}pectral \textbf{A}dversarial \textbf{T}raining (SAT), a simple yet effective adversarial training approach for GNNs. SAT first adopts a low-rank approximation of the graph structure based on spectral decomposition, and then constructs adversarial perturbations in the spectral domain rather than directly manipulating the original graph structure. 
        To investigate its effectiveness, we employ SAT on three widely used GNNs. Experimental results on four public graph datasets demonstrate that SAT significantly improves the robustness of GNNs against adversarial attacks without sacrificing classification accuracy and training efficiency.
    \end{abstract}

    }

\maketitle

\IEEEdisplaynontitleabstractindextext

%
\IEEEpeerreviewmaketitle

\IEEEraisesectionheading{\section{Introduction}}
\IEEEPARstart{G}{raph} Neural Networks (GNNs)~\cite{gnn_survey}, as powerful architectures for modeling graph-structured data, have attracted enormous attention in both academia and industry in recent years. Unlike Convolutional Neural Networks (CNNs) that operate on Euclidean data, GNNs provide an effective and efficient framework, namely \textit{message passing and aggregation}, allowing to generalize the CNN architectures to non-Euclidean domains (graphs and manifolds), and learn the relationships (edges) between a set of objects (nodes). Due to their strong representation ability, GNNs have achieved significant success in a variety of graph applications, such as social networks~\cite{fan2019graph}, knowledge graphs\cite{zhang2020relational}, recommender systems~\cite{liu2020modelling}, and even life science~\cite{li2021dgl}.

Nevertheless, recent efforts show that neural networks are vulnerable to adversarially designed inputs, \emph{i.e.}, \emph{adversarial examples}~\cite{goodfellow2014explaining}, and naturally, GNNs inherit such defects. Despite the promising advancement of GNNs, they still suffer from potential security threats in many graph-related tasks.  Specifically, a well-trained GNN can be easily fooled by slight but carefully designed perturbations on the graph structure or node features, which try to mislead the model by malicious input instances. For example, attackers may modify (add or remove) a small number of connections (edges), such that the target node could be misclassified \cite{zugner2018adversarial,DBLP:conf/icml/DaiLTHWZS18} or the overall model performance would have a significant degradation \cite{DBLP:conf/iclr/Daniel19}. How to resist such adversarial behaviors has so far become a crucial open problem, which motivates researchers to develop more effective methods to enhance the robustness of GNNs.

\begin{figure}[t]
    \centering
    \includegraphics[width=\linewidth]{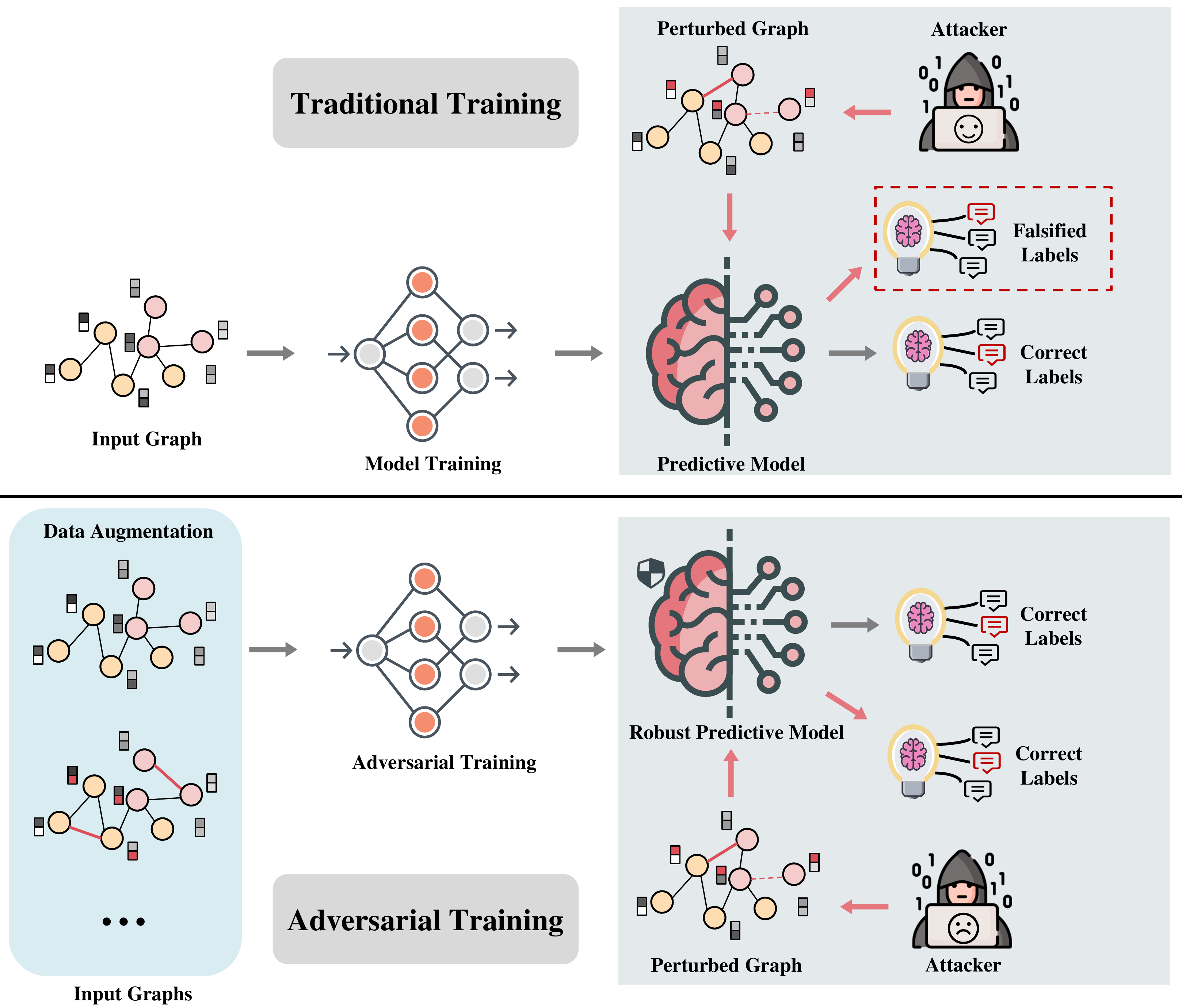}
    \caption{Illustration of traditional training compared with adversarial training.}
    \label{fig:at}
\end{figure}

\textit{Adversarial Training} (AT) has recently emerged as a general approach to improve the resistance of GNNs against adversarial attacks. As shown in Figure~\ref{fig:at}, AT simulates the adversarial attacks during the training phase and encourages the smoothness of prediction outputs. It has been proven as an effective data augmentation and regularization technique to enhance the robustness against adversarial examples \cite{DBLP:conf/iclr/KurakinGB17,DBLP:conf/iclr/MiyatoDG17,he2018adversarial}. Current AT approaches on graph data are typically motivated by the vision research that adds imperceptible perturbations on image data~\cite{goodfellow2014explaining}. These existing methods can be summarized into two generic categories: (i) \textbf{structure-based AT}. The graph topology itself is subject to attacks by adversarially injecting or removing edges or nodes, and hence the adversarial examples can likewise be constructed by such modifications.  In this way, Xu \emph{et al.}~\cite{xu2019topology} present the first optimization-based AT, where the adversarial input graph is formed by a PGD attack~\cite{xu2019topology} that greedily leverages the gradients and a projection operation to sequentially modify the nodes' connections.  (ii) \textbf{feature-based AT}. It is challenging to facilitate the difficulty of tackling graph data since one has to maintain their properties, \emph{e.g.}, discreteness\footnote{Generally, if we use an adjacency matrix to denote a simple unweighted graph, any values would be meaningless except 0 and 1.}. Therefore, existing works mainly manipulate the node input features or hidden representations with continuous perturbations to avoid directly manipulating graph structures. As such, Feng \emph{et al.}~\cite{feng2019graph} dynamically apply graph perturbations on input node features and encourage the smoothness of prediction outputs via virtual adversarial training. Jin \emph{et al.}~\cite{jin2019latent} propose latent adversarial training, which constructs adversarial examples by perturbing the latent representations instead of directly perturbing the graph.

Although these works attempt to enhance the robustness of GNNs by employing adversarial training on the graph domain, there still exist several problems to be addressed:  (i) For structure-based AT, an immediate obstacle arises from computational overhead. Existing works on generating adversarial examples by structural attacks, \emph{e.g.}, \cite{zugner2018adversarial} and \cite{zugner2019certifiable}, usually require substantially more costs than standard attacks. Besides, the resulting model jointly trained with a specific attack may not easily generalize to different attacks that have never been seen before. In conclusion, directly applying AT to discrete graph data is inefficient and fraught with challenges. (ii) For feature-based AT, it treats node input features or latent representations as independent of each other, including connected instances. The perturbations are performed on the feature space separately while the relationship between instances is discarded, which would fail to capture the structural information of a graph and generalize to structural attacks. However, it has been proven that structural attack is more powerful than feature attack~\cite{abs-2005-11560}. Hence, standard adversarial training on feature space is not sufficient to resist stronger structural adversarial attacks (we provide detailed experimental evidence in Section \ref{sec:exp2}).

In light of these challenges, in this work, we propose a novel \textit{Spectral Adversarial Training (SAT)} method to learn robust GNNs against adversarial attacks. SAT enjoys the advantages of structure and feature-based AT, as well as overcoming both of these obstacles based on the matrix perturbation theory. Specifically, SAT first forms a low-rank approximation of the graph structure by spectral decomposition. Since the spectrum characterizes the topological property of the graph, SAT approximates the structure perturbations in terms of the change of the spectrum to facilitate adversarial training. Then, we perform perturbations on the spectrum by treating the eigenvalues and eigenvectors (eigenpairs) as node structural features. The standard AT is applied on the eigenpairs to simulate the structure-based AT in a continuous space. In this regard, SAT additionally incorporates two spectral adversarial training regularizers into the objective function during training, for the sake of enforcing the model to yield a smoothed output even when the adversarial attacks are presented.

The main contributions of this paper are summarized as follows:
\begin{itemize}
    \item We thoroughly investigate related works and subsequently give a unified formulation of AT on graph data. We systematically categorize existing works related to graph-based AT into structure-based and feature-based methods and shed light on the obstacles to overcome.
    \item We propose a novel approach \textit{Spectral Adversarial Training  (SAT)}, which approximates the low-rank graph by spectral decomposition and employs adversarial training on the graph spectrum to regularize GNN models and improve their robustness against adversarial attacks.
    \item We demonstrate the robustness and generalization ability of SAT employed on a line of widely used GNNs. Comprehensive experiments on four node classification datasets validate that SAT is capable to improve the model robustness without sacrificing classification power and training efficiency.
\end{itemize}

The rest of this article is organized as follows. In Section \ref{sec:related_work}, we summarize the related works of GNNs and adversarial defense on graph data. In Section \ref{sec:preliminary}, we give notations and preliminaries. In Section \ref{sec:method} and Section \ref{sec:exp}, we illustrate our proposed method and present the experimental results with sufficient analysis, respectively. Finally, we draw our conclusion and highlight future works in Section \ref{sec:conclusion}.

\section{Related Works}\label{sec:related_work}
In this section, we will briefly review the recent advances regarding the
GNN model family, and then describe the literature on graph adversarial defense research closely related to this work.

\subsection{Graph Neural Networks}
GNNs~\cite{scarselli2009graph,gnn_survey} are a class of deep learning models designed to learn representations of data described by graphs. They extend the traditional neural networks to directly operate on the graph-structured data, and provide an elegant way to implement node-level, edge-level, and graph-level prediction tasks~\cite{DBLP:journals/tnn/WuPCLZY21}.

The key to the success of GNNs is the message passing scheme between the nodes and their neighborhoods (\emph{a.k.a.} neighborhood aggregation), which recursively aggregates the messages (embeddings) based on the adjacency structure of the graph. Following this scheme, Graph Convolutional Networks (GCNs)~\cite{KipfW17} are proposed to apply convolutional operations on the local neighborhoods of the nodes. GCNs capture the graph structure and also borrow feature information to learn an embedding for each node that contains neighboring information. Wu \emph{et al.}~\cite{WuSZFYW19} propose Simple Graph Convolution (SGC) that simplifies GCNs by removing nonlinearities and applying the $K$-th power of graph convolution in a single layer. In a following work, 
Zhu \emph{et al.}~\cite{zhu2021simple} propose Simple Spectral Graph convolution (S$^2$GC) that generalizes SGC to deeper architectures while limiting over-smoothing issues. For a comprehensive overview we refer readers to a survey~\cite{DBLP:journals/tnn/WuPCLZY21} on graph neural networks.

Due to the excellent representation ability, GNN-based models have achieved advanced performance on several tasks and play a vital role in machine learning fields. However, recent studies show that GNNs are vulnerable to adversarial attacks~\cite{zugner2018adversarial,zugner2019certifiable,zhu2021simple}, particularly deliberate perturbations on the graph structures. The robustness of GNNs has drawn significant attention ever since, which motivates us to design a robust training approach in this work.

\subsection{Adversarial Defense on Graph Data}
Extensive efforts so far have been dedicated to the design of defense strategies that improve the robustness of GNNs and resist adversarial attacks. They can be generally categorized into three classes~\cite{sun2018adversarial}: adversarial purification, robust architectures, and adversarial training. As the focus of our work is on adversarial training, we will draw more attention to it in the following. We refer readers to \cite{sun2018adversarial,chen2020survey} and the references therein for more thorough reviews of adversarial defenses on graphs.

\subsubsection{Adversarial Purification}
To reduce the effect of adversarial behaviors, a straightforward way of defense is to detect and remove potential perturbations in a graph. Based on this insight, Wu \emph{et al.}~\cite{wu2019adversarial} propose to inspect the graph and recover the suspicious connections via Jaccard similarity scores of features between connected nodes. Xu \emph{et al.}~\cite{xu2019characterizing} propose a different approach to detect malicious edges with a filter-and-sample framework. Beyond these edge-based methods, Entezari \emph{et al.}~\cite{entezari2020all} propose to use Singular Value Decomposition (SVD) to form a low-rank approximation of graph adjacency matrix, and thus reduce the adversarial effects.

\subsubsection{Robust Architectures}
Rather than directly inspecting poisoned graph structures, such as edges or nodes before inputting to the models, several works attempt to design more robust architectures for GNNs. Specifically,  Zhu \emph{et al.}~\cite{zhu2021simple} adopt Gaussian distributions as node representations and employ attention mechanisms to dynamically adjust the propagation of adversarial attacks in GNNs. As the vulnerability of GNNs is mostly attributed to the non-robust aggregation scheme~\cite{median}, Jin \emph{et al.}~\cite{JinDW0LT21} propose Similarity Preserving GCN (SimPGCN) that preserves node similarly and adaptively capture the graph structure during the aggregation process. Chen \emph{et al.}~\cite{median} adopt robust aggregation schemes, median and trimmed mean, instead of weighted mean to enhance the robustness of GNNs.

\subsubsection{Adversarial Training}
While adversarial perturbations are initially used to construct adversarial examples and fool the neural networks, the same sword can be utilized by defenders to improve the robustness of their models against adversarial attacks~\cite{sun2018adversarial}. Adversarial training (AT) on graph data is inspired by research in computer vision, where the perturbations are constructed in continuous input domains. So far, a variety of works~\cite{feng2019graph,deng2019batch,JinZ20} employ virtual adversarial training on node features, to enforce the smoothness of the models' output distributions against adversarial inputs. In particular, Deng \emph{et al.}~\cite{deng2019batch} propose batch virtual AT to smooth the outputs of GNNs, Feng \emph{et al.}~\cite{feng2019graph} devise a dynamic regularizer encouraging GNN models to learn representations that are robust to perturbations. In the ensuing work, Hu \emph{et al.}~\cite{hu2021robust} design neighbor perturbations to restrict the perturbation directions on node features and prevent the propagation of perturbed neighbors. Jin \emph{et al.}~\cite{JinZ20} apply AT on the latent representations of the nodes to improve the robustness and generalization ability of GNNs.

The aforementioned works are typically feature-based AT. However, the structural information has not been fully exploited during training. The structure-based AT perturbs the graph by adding or dropping edges or nodes, which is also an effective way to defend against attacks. Dai \emph{et al.}~\cite{DBLP:conf/icml/DaiLTHWZS18} propose to randomly drop edges during each training step. Such a simple strategy has shown its effectiveness in improving the GNNs' robustness. One step further, Xu \emph{et al.}~\cite{xu2019topology} design an optimization-based AT, where adversarial edges are generated by a topology attack. The presented work has demonstrated the effectiveness of employing structural AT on graph data. To facilitate adversarial training via practical adversarial examples, a growing number of researchers~\cite{abs-1903-05994,xu2019topology} seek to employ attack-oriented perturbations, which are generated based on existing network adversarial attacks of PGD~\cite{xu2019topology} and Metattack~\cite{DBLP:conf/iclr/Daniel19}.

Although these works attempt to handle similar problems, \emph{i.e.}, improving the robustness of GNNs via practical adversarial training, they either require substantial costs to construct adversarial examples in the discrete graph domain (structure-based AT) or discard the graph topology that is crucial for GNNs (feature-based AT). In light of these challenges, our work attempts to employ AT on the graph spectrum. Specifically, it is able to approximate the structure perturbations in terms of the changes of the eigenpairs. Thus, the discrete optimization problem is transformed into a continuous one and the computational overhead is greatly alleviated.

\section{Preliminary}\label{sec:preliminary}
In this section, we first give some notations and review the family of GNN models, and then formulate the problem of AT on graph data. For convenience, our frequently used notations and formulas are summarized in Table \ref{table:notation}.
\begin{table}[h]
    \centering
    \caption{Frequently used notations in this paper.}
    \label{notation}
    \resizebox{\linewidth}{!}{
        \begin{tabular}{cl}
            \toprule
            Notations                  &
            Descriptions                 \\

            \midrule

            \midrule
            \begin{tabular}[c]{@{}c@{}} $V$ \end{tabular}  &
            \begin{tabular}[c]{@{}c@{}} Set of nodes in the graph \end{tabular}    \\

            \midrule
            \begin{tabular}[c]{@{}c@{}} $E$ \end{tabular}  &
            \begin{tabular}[c]{@{}c@{}} Set of edges in the graph \end{tabular}    \\

            \midrule
            \begin{tabular}[c]{@{}c@{}} $G$ \end{tabular}  &
            \begin{tabular}[c]{@{}c@{}} Graph representation of the data \end{tabular}   \\

            \midrule
            \begin{tabular}[c]{@{}c@{}} $Y$ \end{tabular} &
            \begin{tabular}[c]{@{}c@{}} Set of class labels of the nodes \end{tabular}   \\

            \midrule
            \begin{tabular}[c]{@{}c@{}} $n$, $c$, $d$ \end{tabular} &
            \begin{tabular}[c]{@{}c@{}} Number of the nodes, classes and feature dimensions \end{tabular}   \\

            \midrule
            \begin{tabular}[c]{@{}c@{}}  $\mathbf{A}$ \end{tabular} &
            \begin{tabular}[c]{@{}c@{}} Adjacency matrix of the graph \end{tabular}
            \\

            \midrule
            \begin{tabular}[c]{@{}c@{}}  $\hat{\mathbf{A}}$  \end{tabular} &
            \begin{tabular}[c]{@{}c@{}} Normalized Laplacian matrix of the graph \end{tabular}
            \\

            \midrule
            \begin{tabular}[c]{@{}c@{}}  $\mathbf{X}$ \end{tabular} &
            \begin{tabular}[c]{@{}c@{}} Feature matrix of the nodes\end{tabular}
            \\

            \midrule
            \begin{tabular}[c]{@{}c@{}}  $\mathbf{D}$ \end{tabular} &
            \begin{tabular}[c]{@{}c@{}} Diagonal degree matrix of the nodes\end{tabular}
            \\

            \midrule
            \begin{tabular}[c]{@{}c@{}}  $f_\Theta$ \end{tabular} &
            \begin{tabular}[c]{@{}c@{}} Graph Neural Network model parameterized by $\Theta$\end{tabular}
            \\

            \midrule
            \begin{tabular}[c]{@{}c@{}}  $\mathbf{W}$ \end{tabular} &
            \begin{tabular}[c]{@{}c@{}} Trainable weight matrix \end{tabular}
            \\

            \midrule
            \begin{tabular}[c]{@{}c@{}}  $\mathbf{Z}_{\text{GCN}}$, $\mathbf{Z}_{\text{SGC}}$, $\mathbf{Z}_{\text{S}^2\text{GC}}$ \end{tabular} &
            \begin{tabular}[c]{@{}c@{}} Prediction outputs of GCN, SGC and S$^2$GC \end{tabular}
            \\

            \midrule
            \begin{tabular}[c]{@{}c@{}}  $\mathbf{H}^{(k)}$ \end{tabular} &
            \begin{tabular}[c]{@{}c@{}} Hidden output of layer $k$\end{tabular}
            \\
            \midrule
            \begin{tabular}[c]{@{}c@{}}  $r$\end{tabular} &
            \begin{tabular}[c]{@{}c@{}} Approximated rank of the graph structure \end{tabular}
            \\

            \midrule
            \begin{tabular}[c]{@{}c@{}}  $\delta_f$, $\delta_l$,\\ $\delta_s$, $\delta_\mathbf{U}$, $\delta_\mathbf{\Lambda}$\end{tabular} &
            \begin{tabular}[c]{@{}c@{}} Perturbations on node features, latent representations, \\graph structures, eigenvectors and eigenvalues \end{tabular}
            \\

            \midrule
            \begin{tabular}[c]{@{}c@{}}  $\Phi$($\cdot$) \end{tabular} &
            \begin{tabular}[c]{@{}c@{}} Perturbation space of a specific instance \end{tabular}
            \\

            \midrule
            \begin{tabular}[c]{@{}c@{}}  $\mathcal{L}_{\text{train}}$ \end{tabular} &
            \begin{tabular}[c]{@{}c@{}} Training loss of GNNs \end{tabular}
            \\
            \midrule
            \begin{tabular}[c]{@{}c@{}}  $\lambda$, $u$ \end{tabular} &
            \begin{tabular}[c]{@{}c@{}} Eigenvalue and the associated eigenvector \end{tabular}
            \\

            \midrule
            \begin{tabular}[c]{@{}c@{}}
            $\mathbf{\Lambda}_r = diag(\lambda_1, \lambda_2, ..., \lambda_r)$,\\$\mathbf{U}_r=[u_1, u_2, ..., u_r]$
            \end{tabular} &
            \begin{tabular}[c]{@{}c@{}}  Top-$r$ largest eigenvalues and corresponding eigenvectors\end{tabular}
            \\

            \bottomrule
        \end{tabular}
    }
    \label{table:notation}
\end{table}

\subsection{Notations}
Consider an undirected attributed graph $G = (\mathbf{A}, \mathbf{X})$ that has $n$ nodes, with $\mathbf{A} \in \{0,1\}^{n\times n}$ denoting the adjacency matrix and $\mathbf{X} \in \mathbb{R}^{n\times d}$ representing the node feature matrix, where $d$ is the dimensionality. Without loss of generality, we focus on an attributed, undirected graph in this paper. Formally, we denote the set of nodes as $V=\{v_i\}$ and the set of edges as $E\subseteq V\times V$, respectively. In the context of semi-supervised node classification, each node $v_i$ is associated with a corresponding node label $y_i\in Y=\{0,1,\dots,c-1\}$, where $c$ is the number of class labels.

\subsection{Graph Neural Network Family}\label{sec:gnn}
\subsubsection{Graph Convolutional Network (GCN)}
Graph Convolutional Network (GCN)~\cite{KipfW17} is one of the classical models in graph-based learning, which has been broadly used in many tasks. The purpose of the GCN is to leverage the geometrical properties into the neural network by an iterative message aggregation scheme. Under the setting of semi-supervised node classification, the $(l + 1)$-th hidden output of GCN can be formulated as follows:
\begin{equation}\label{eq:hidden}
    \mathbf{H}^{(l+1)} = \sigma(\mathbf{\widetilde{D}}^{-\frac{1}{2}}\mathbf{\widetilde{A}}\mathbf{\widetilde{D}}^{-\frac{1}{2}}\mathbf{H}^{(l)}\mathbf{W}^{(l)}).
\end{equation}
Here, $\mathbf{\widetilde{A}} = \mathbf{A} + \mathbf{I}$ is the adjacency matrix of the graph $G$ with self-connections, $\mathbf{I}\in \mathbb{R}^{n\times n}$ is the identity matrix, and the $i$-th diagonal element of $\mathbf{\widetilde{D}}$, $\mathbf{\widetilde{D}}_{ii} = \sum_{j}\mathbf{\widetilde{A}}_{ij}$, is degree of node $i$ plus 1. $\mathbf{W}^{(l)}\in \mathbb{R}^{d\times d^{\prime}}$ is the trainable weight matrix of layer $l$ which learns useful features and transforms the embedding size from $d$ to $d^{\prime}$. Here we omit the bias terms in each layer for the sake of simplicity. $\sigma(\cdot)$ denotes the activation function, such as ReLU~\cite{glorot2011deep} in this paper. Initially, $\mathbf{H}^{(0)} = \mathbf{X}$. In particular, a two-layer GCN with ReLU as hidden activation and softmax as output is widely adopted. Thus the semi-supervised classification model can be described as:
\begin{equation}\label{eq:gcn}
    \mathbf{Z}_{\text{GCN}} = \mathrm{softmax}(\mathbf{\hat{{A}}} \cdot \text{ReLU}(\mathbf{\hat{{A}}} \mathbf{X} \mathbf{W^{(0)})} \mathbf{W^{(1)}}),
\end{equation}
where $\mathbf{\hat{{A}}} = \mathbf{\widetilde{D}}^{-\frac{1}{2}}\mathbf{\widetilde{A}}\mathbf{\widetilde{D}}^{-\frac{1}{2}}$ is the normalized Laplacian matrix of graph $G$.

\subsubsection{Simple Graph Convolution (SGC)}
The vanilla GCN recursively but repeatedly computes the aggregated messages from the nodes' neighborhoods, resulting in excessive computation during training. To reduce the computation overhead, Wu \emph{et al.}~\cite{WuSZFYW19} theoretically analyze the architecture of GCN and propose to linearize it. Specifically, the resulting linear model, SGC, removes the nonlinear activation function in the hidden layers and collapses the weight matrices between consecutive layers. Therefore, the output of the GCN is simplified as:
\begin{equation}\label{eq:sgc}
    \mathbf{Z}_{\text{SGC}} = \mathrm{softmax}(\mathbf{\hat{A}}^K\mathbf{X}\mathbf{W}),
\end{equation}
where $\mathbf{W}=\mathbf{W}^{(0)} \mathbf{W}^{(1)}$ is the redefined weight matrix, and $K$ is the number of hidden layers and often set as 2. SGC captures structural information from the $K$-hop neighborhoods of the nodes in the graph, by simply applying the $K$-th power of the transition matrix $\mathbf{\hat{A}}$ in a single neural network layer. It is worth mentioning that the term of $\mathbf{\hat{A}}^K \mathbf{X}$ is usually a constant and can be pre-computed, which reduces the computation overhead during training.

\subsubsection{Simple Spectral Graph Convolution ($\text{S}^2$GC)}
Inspired by the previous works~\cite{WuSZFYW19,KlicperaBG19}, $\text{S}^2$GC~\cite{zhu2021simple} has been recently proposed as a state-of-the-art model. Similar to SGC, $\text{S}^2$GC is a linear learner which seeks to address the issues of over-smoothing and high computational costs encountered in the literature. $\text{S}^2$GC adopts a simple but effective graph filter that aggregates the neighborhoods with various sizes, and is beneficial for capturing the global and local information of each node in the graph. Typically, $\text{S}^2$GC generalizes the output of SGC as:
\begin{equation}\label{eq:ssgc}
    \mathbf{Z}_{\text{S}^2\text{GC}}=\mathrm{softmax}\left(\frac{1}{K} \sum_{k=1}^{K}\left((1-a) \mathbf{\hat{{A}}}^{k} \mathbf{X}+a \mathbf{X}\right) \mathbf{W}\right),
\end{equation}
where $a\in [0,1]$ is a scalar to balance the aggregations of the nodes and their neighborhoods. Similar to SGC, the graph filter can be also pre-computed and the resulting model represents a good trade-off between classification accuracy and computation overhead.

\subsection{Adversarial Training Formulation}
AT~\cite{goodfellow2014explaining} is a regularization technique for standard training and has been demonstrated effective to improve the model robustness against adversarial attacks. Generally, AT can be viewed as a min-max game: a network defender seeks to minimize the training loss on the graph, while the adversary constructs adversarial examples to maximize the training loss. Through alternating the two opposite (minimization and maximization) processes iteratively, the learned model is expected to have better robustness and generalization ability, not only against benign examples that have not yet been seen in the training data, but also against adversarial examples. We extend the formulation of AT in ~\cite{jinsurvey}, to make it a unified one in the graph setting:
\begin{equation}\label{eq:at}
    \min _{\Theta} \max _{\hat{G} \in \Phi(G) \atop \mathbf{\hat{H}}^{(k)} \in \Phi(\mathbf{H}^{(k)}) } \sum_{v_i\in V_L} \mathcal{L}_{\text{train}}(f_{\Theta} (\hat{G}; \mathbf{\hat{H}}^{(k)}), y_{i}), \quad k\in \{0,1\}.
\end{equation}
Here $\hat{G} \in \Phi(G)$ is the input graph which can be either a well-designed graph to achieve the goal of AT or a clean graph without perturbations. $\mathbf{\hat{H}}^{(k)}$ is the perturbed or unperturbed feature representations of the $k$-th layer. In most cases, the perturbations are added on the input features or the first hidden output~\cite{JinZ20}, \emph{i.e.}, $\mathbf{H}^{(0)}$ ($\mathbf{X})$ or $\mathbf{H}^{(1)}$. $\Phi$ denotes the perturbation function of adversarial examples \emph{w.r.t.} the given inputs. $f_\Theta$ is a GNN model parameterized by $\Theta$ and takes a graph as input. $V_L$ is the set of labeled nodes for training, while $v_i$ and $y_i$ are the labeled node in $V_L$ and the corresponding class label, respectively. $\mathcal{L}_{\text{train}}$ is the training loss function, which can be formulated as follows:
\begin{equation}\label{eq:loss_train}
    \mathcal{L}_{\text{train}} = \Omega + \gamma\Gamma,
\end{equation}
where $\Omega$ is the classification loss that measures the divergence between the prediction and the ground-truth.  $\Gamma$ is an additional regularization term, \emph{e.g.}, $\ell_2$ regularization on $\Theta$. $\gamma\geq 0$ is the parameter balancing the two terms.

According to Equation \eqref{eq:at}, the adversarial perturbations can be added on the input graph, including the adjacency structures $\mathbf{A}$ and the node features $\mathbf{X}$, or the hidden representation $\mathbf{H}^{(k)}$. Since the hidden representation $\mathbf{H}^{(k)}$ can be treated as the transformed node features, we group AT with perturbations on $\mathbf{X}$ and $\mathbf{H}^{(k)}$ as feature-based AT, and the remaining as structure-based AT.

\subsubsection{Structure-based AT Formulation}
The structure-based AT mainly focuses on the graph structure denoted by $\mathbf{A}$. Unlike on non-graph structured data, \emph{e.g.}, image or text data, AT cannot be directly employed on the graph structure due to its discreteness. Naturally, there must be constraints on the adversarial perturbations, which ensure the discreteness of the graph structure. Specifically, we define the structure-based AT on top of Equation \eqref{eq:at}, as:
\begin{equation}\label{eq:at_structure}
    \begin{aligned}
         & \min _{\Theta} \max _{\delta_s } \sum_{v_i\in V_L} \mathcal{L}_{\text{train}}(f_{\Theta} (\mathbf{A}+\delta_s, \mathbf{X}), y_{i}), \\
         & \text{s.t.} \quad \mathbf{A}+\delta_s \in \{0,1\}^{n\times n},
    \end{aligned}
\end{equation}
where $\delta_s$ denote the \underline{s}tructural perturbations.

Although generating adversarial examples based on randomly dropping is quite cheap to implement, there is little performance gain against adversarial attacks with such a strategy~\cite{jinsurvey}. Currently, the mainstream structure-based AT methods are attack-oriented, that is, the adversarial perturbations are generated by powerful attackers. However, adversarial attacks on graph data are frequently insufficient, and generating such adversarial examples comes at a high cost, which limits their usability.

\subsubsection{Feature-based AT Formulation}
The feature-based AT generates adversarial examples in a different way. The perturbations are added on the node features, either the original input features or the transformed ones in the latent space. Similar to structure-based AT, we define the feature-based AT as:
\begin{equation}\label{eq:at_feature}
    \min _{\Theta} \max _{\delta_f, \delta_l } \sum_{v_i\in V_L} \mathcal{L}_{\text{train}}(f_{\Theta} (\mathbf{A}, \mathbf{X}+\delta_f;\mathbf{H}^{(1)}+\delta_l), y_{i}),
\end{equation}
where $\delta_f$ and $\delta_l$ are the perturbations added on the node input \underline{f}eature and the \underline{l}atent representations, respectively. Because the features are often in the continuous domain, the generated perturbations can have any values and are confined by a tiny constant.

Existing feature-based AT methods, however, assume that the nodes are independently distributed across the graph and the relationships between the samples are less considered or even disregarded. As a result, they often fail to facilitate AT on the graph structure and are insufficient to withstand more powerful adversarial attacks, particularly structure-based ones (See Section~\ref{sec:exp2}).

\section{Methodology}\label{sec:method}
In this section, we first introduce the matrix perturbation theory that lays as the foundation of our work, followed by the low-rank approximation of the graph structure. We then outline how we construct adversarial examples based on spectral perturbations, and present the overall objective function and the algorithm for Spectral Adversarial Training (SAT). Finally, we analyze the time and memory complexity of SAT.

\subsection{Matrix Perturbation Theory}\label{sec:mp_theory}
In the area of complex network analysis, one of the most important topics is to study the robustness of a graph \cite{milanese2010approximating,yan2014eigenvector,yang2020abstract}. The robustness of a graph is evaluated by certain metrics under the perturbations in terms of both structural and spectral changes. The structural perturbations include deletion or addition of edges or nodes, and modification of the edge weights; the spectral changes consist of perturbations on the eigenvalues and eigenvectors of the adjacency or Laplacian matrix. There is an intriguing problem to investigate: \emph{how does a slight change in the input affects the output?}

Formally, given a symmetric matrix $\mathbf{A}$ and some arbitrary function $\phi$ which operates on $\mathbf{A}$, we are interested in understanding how a slight perturbation added to the matrix affects the behavior of $\phi$. Noticing that a low dimensional representation of $\mathbf{A}$ can be computed using spectral decomposition, we are able to estimate the changes from the spectrum perspective, \emph{i.e.}, the eigenpairs of $\mathbf{A}$.
Specifically, we are given a symmetric $n\times n$ matrix $\mathbf{A}$ with its $r$ dominant eigenvalues $\lambda_1 \geq \lambda_2 \geq  \dots \geq \lambda_r$ and the corresponding eigenvectors $u_1, u_2, \dots, u_r$, respectively. If the matrix is modified from $\mathbf{A}$ to $\mathbf{A} + \Delta_\mathbf{A}$ where the perturbation $||\Delta_\mathbf{A}||$ is sufficiently small, we look for $\Delta\lambda_{i}$ and $\Delta u_{i}$ that satisfy the equation:
\begin{equation}
    (\mathbf{A}+\Delta_\mathbf{A})(u_{i}+\Delta u_{i})=(\lambda_{i}+\Delta \lambda_{i})(u_{i}+\Delta u_{i}),
\end{equation}
for $i=1, 2, \dots, r$.

Intuitively, it is straightforward to recompute the perturbed eigenpairs of $\mathbf{A}$ directly based on spectral decomposition. However, finding the eigenpairs results in high computation overhead when $\mathbf{A}$ is high-dimensional. Given that the perturbation on $\mathbf{A}$ is sufficiently small, we wish to estimate the perturbed eigenpairs of $\mathbf{A} + \Delta_\mathbf{A}$ based on $\mathbf{A}$ and its eigenpairs calculated aforehand. According to the matrix perturbation theory \cite{stewart1990matrix},  the changes of eigenvalues and eigenvectors can be estimated as follows:
\begin{equation}\label{eq:change}
    \begin{aligned}
        \Delta\lambda_{i} &= \hat{\lambda}_i  - \lambda_i \approx u_{i}^{T}\Delta_\mathbf{A} u_{i},                                    \\
        \Delta u_{i} &= \hat{u}_i - u_i     \approx \sum_{r\neq i}\frac{u_{r}^{T}\Delta_\mathbf{A} u_{r}}{\lambda_i - \lambda_r + \lambda_{\epsilon}}u_r,
    \end{aligned}
\end{equation}
where $\hat{u}_i$ and $\hat{\lambda}_i$ are the $i$-th perturbed eigenvector and eigenvalue of $\mathbf{A} + \Delta_\mathbf{A}$, respectively. $\lambda_{\epsilon}$ is a small constant to avoid the numerical instability.

The matrix perturbation theory shows that the perturbations on $\mathbf{A}$ are directly related to the eigenpairs, and we are able to update the perturbed eigenpairs of $\mathbf{A} + \Delta_\mathbf{A}$ based on $\mathbf{A}$ and its eigenpairs if the changes are sufficiently small. This is the primary motivation that drives us to go deeper into AT from the spectrum perspective. We concentrate on studying the direct perturbation training in the continuous domain with spectral decomposition and propose an adversarial regularization method to improve the model robustness. We construct adversarial perturbations on the spectrum of the graph during the training phase, rather than simply modifying the discrete graph structure or empirically employing other continuous components to generate adversarial examples.

\begin{figure*}[t]
    \centering
    \includegraphics[width=0.9\textwidth]{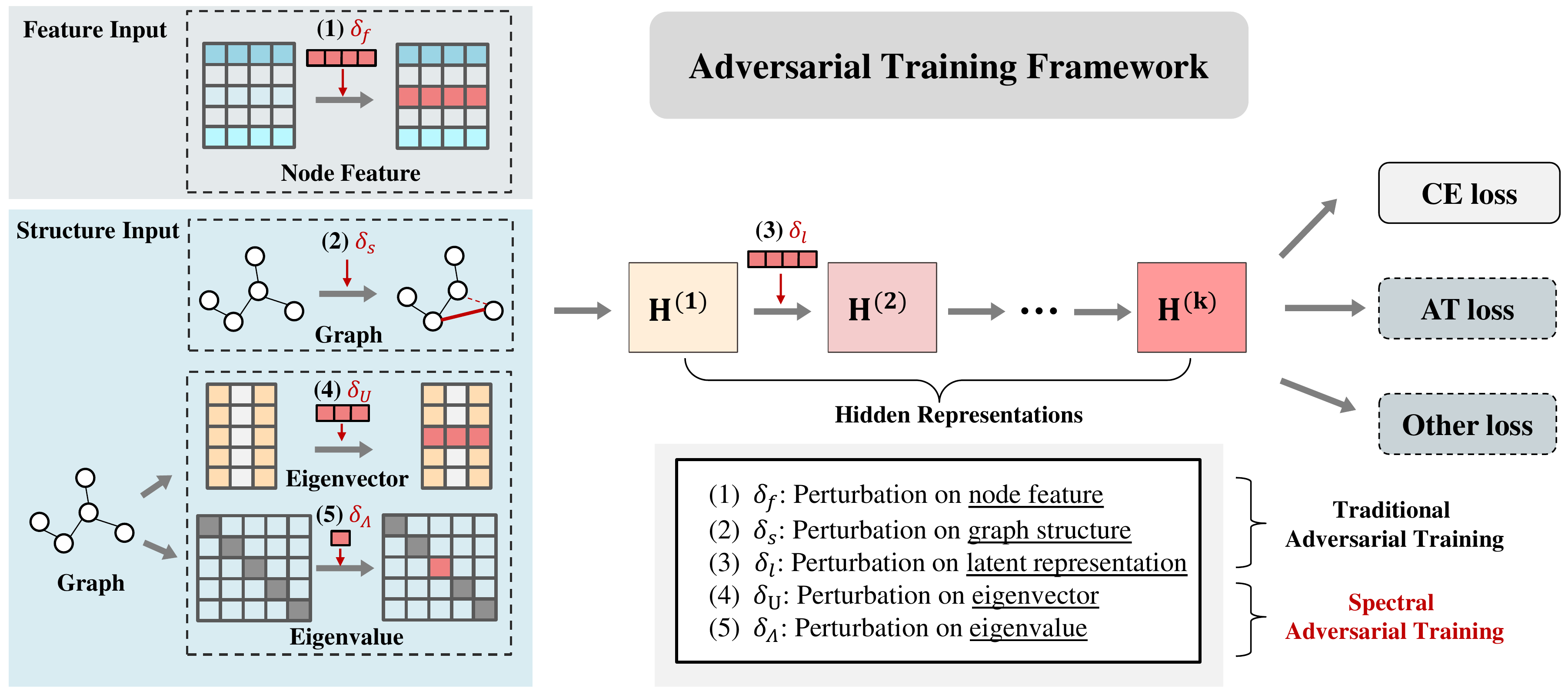}
    \caption{AT framework for various types of perturbations. $\delta_f$ and $\delta_l$ are adversarial perturbations generated for feature-based AT, while $\delta s$ is structure-based AT. Our proposed SAT incorporates the structural information by performing perturbations on the eigenpairs.}
    \label{fig:framework}
\end{figure*}

\subsection{Low-Rank Approximation of Graph Structure}
Several works \cite{goodfellow2014explaining,DBLP:conf/iclr/MiyatoDG17,he2018adversarial} on AT have developed methods to add perturbations on image raw features or embeddings that are continuous. Since the graph structure is discrete, previous methods cannot be directly extended to generate structure perturbations for graph neural networks. Inspired by the matrix perturbation theory, one possible way is to transform the discrete optimization problem into a continuous one that can be solved by well-established AT approaches. As illustrated in Section \ref{sec:mp_theory},  the structural perturbations can be viewed as the changes in the eigenvalues and eigenvectors. This motivates us to approximate the structural perturbations in terms of spectral changes.

Firstly, we observe that the normalized Laplacian matrix $\hat{\mathbf{A}} =\mathbf{\widetilde{D}}^{-\frac{1}{2}}\mathbf{\widetilde{A}}\mathbf{\widetilde{D}}^{-\frac{1}{2}}$ encodes the topological information of the graph $G$. As $\hat{A}$ is a real symmetric semi-definite matrix, it can be decomposed to a set of orthogonal eigenvectors and associated nonnegative eigenvalues. Thus, we are able to modify the obtained eigenvalues and eigenvectors to generate adversarial examples from the continuous domain.

Suppose that $\mathbf{\Lambda}_r = diag(\lambda_1, \lambda_2, ..., \lambda_r)\in \mathbb{R}^{r\times r}$ represents the set of  $r$ dominant eigenvalues and $\mathbf{U}_r=[u_1, u_2, ..., u_r]\in \mathbb{R}^{n\times r}$ denotes the set of associated eigenvectors. Spectral decomposition is an elegant tool to compute the best rank-$r$ approximation of matrix $\hat{\mathbf{A}}$. The rank-$r$ approximation of $\hat{\mathbf{A}}$ is calculated as follows:
\begin{equation}
    \hat{\mathbf{A}}_r =\mathbf{U}_r\mathbf{\Lambda}_r \mathbf{U}_r^\mathsf{T},
\end{equation}
where $\hat{\mathbf{A}}_r$ is the rank-$r$ approximation of $\hat{\mathbf{A}}$ derived from spectral decomposition. As stated by Entezari \emph{et al.}~\cite{entezari2020all}, adversarial attacks, such as NETTACK, tend to affect the high-rank (low-valued) components of the graph. Following this insight, we can use a much smaller $r$ to reconstruct a low-rank approximation of the graph structure, where we only choose the $r$ largest eigenvalues and the eigenvectors to project the graph into a low-rank subspace. In addition, using the top-$r$ largest eigenvalues can ensure that they are non-zero and not close to each other, thus avoiding the numerical instability in Equation~\eqref{eq:change}.

Based on the rank-$r$ approximation of $\hat{\mathbf{A}}_r$,  we can reformulate the output of GNNs, by replacing the normalized Laplacian matrix $\hat{\mathbf{A}}$ with $\hat{\mathbf{A}}_r$. In particular, the outputs of SGC and S$^2$GC are respectively written as:
\begin{equation}\label{eq:sgc2}
    \begin{aligned}
        \mathbf{Z}_{\text{SGC}}          & =     \mathrm{softmax}(\mathbf{U}_r\mathbf{\Lambda}_r^K \mathbf{U}_r^\mathsf{T}\mathbf{X}\mathbf{W}),                                             \\
        \mathbf{Z}_{\text{S}^2\text{GC}} & =\mathrm{softmax}\left(\frac{1}{K} \sum_{k=1}^{K}\left((1-a) \mathbf{U}_r\mathbf{\Lambda}_r^K \mathbf{U}_r^\mathsf{T} \mathbf{X}+a \mathbf{X}\right) \mathbf{W}\right),
    \end{aligned}
\end{equation}
where we simplify the computation of $\hat{\mathbf{A}}^K$ into  $\mathbf{U}_r\mathbf{\Lambda}_r^K \mathbf{U}_r^\mathsf{T}$ using the properties of eigenpairs of a symmetric matrix.

Observe that the outputs of SGC and S$^2$GC are now substantially easier to compute. One can easily compute $\mathbf{U}_r\mathbf{\Lambda}_r^K \mathbf{U}_r^\mathsf{T}$ even $K$ is larger. It is worth noting that since the attackers tend to perturb the high-rank components of the graph spectrum, this simplification does not negatively affect the performance. Instead, it benefits to improve the robustness of GNNs and effectively reduce the computation overhead.

\subsection{Generating Adversarial Examples}
A straightforward way to generate the adversarial examples is applying gradient ascent on the inputs. Specifically, we treat the eigenvalues and eigenvectors as hyperparameters to compute their gradients and construct the perturbations in terms of $\mathbf{U}$ and $\mathbf{\Lambda}$ independently\footnote{We omit the subscript $r$ in the rest of the paper for brevity.}. Let $\delta=\{\delta_{\mathbf{\Lambda}}, \delta_{\mathbf{U}}\}$ denote the two types of perturbations, representing the changes on  $\mathbf{\Lambda}_r$ and $\mathbf{U}_r$, respectively. Based on Equation \eqref{eq:at}, the objective function of spectral AT could be defined as:
\begin{equation}\label{eq:at_sat}
    \begin{aligned}
         & \min _{\Theta} \max _{\delta } \sum_{v_i\in V_L} \mathcal{L}_{\text{train}}(f_{\Theta} (\mathbf{U}+\delta_\mathbf{U}, \mathbf{\Lambda}+\delta_\mathbf{\Lambda}, \mathbf{X}), y_{i}).
    \end{aligned}
\end{equation}

Although our proposed SAT focuses on the graph structure, it generates adversarial examples from a continuous perspective and can avoid directly tackling the graph structure. We devise two types of spectral adversarial training employed on the eigenvalues and the eigenvectors, respectively. The resulting adversarial loss $\mathcal{L}_{\text{SAT}}^{\text{val}}$ and $\mathcal{L}_{\text{SAT}}^{\text{vec}}$ can be formulated as follows:
\begin{equation}\label{eq:adv}
    \begin{aligned}
        \mathcal{L}_{\text{SAT}}^{\text{vec}} & = \Omega[Q(y|x), P(y|x, \mathbf{U} + \delta^{*}_\mathbf{U}, \mathbf{\Lambda})],            \\
        \mathcal{L}_{\text{SAT}}^{\text{val}} & = \Omega[Q(y|x), P(y|x, \mathbf{U}, \mathbf{\Lambda}+ \delta^{*}_\mathbf{\Lambda}))],      \\
        \text{with} ~ \
        \delta_{\mathbf{U}}^{*}               & = \arg\max_{\delta, ||\delta_{\mathbf{U}}||\leq\varepsilon_1}\Omega[Q(y|x), P(y|x)],       \\
        \delta_{\mathbf{\Lambda}}^{*}         & = \arg\max_{\delta, ||\delta_{\mathbf{\Lambda}}||\leq\varepsilon_2}\Omega[Q(y|x), P(y|x)]. \\
    \end{aligned}
\end{equation}
Here $Q(y|x)$ is the ground-truth distribution of output labels and $P(y|x)=f_\Theta(G)$ denotes the predicted distribution by GNNs. $x$ represents the node features. In this paper, we choose a popular loss function, \emph{i.e.}, cross-entropy (CE) loss, as $\Omega$ for demonstration. Inspired by the linear approximation method~\cite{goodfellow2014explaining} for standard AT, the continuous perturbations $\delta_{\mathbf{\Lambda}}^{*}$ and $\delta_{U}^{*}$ are approximated as follows:
\begin{equation}\label{eq:approx}
    \begin{aligned}
        \delta_{\mathbf{U}}^{*}\approx \varepsilon_1\frac{g}{||g||_{2}},\ \text{where}\ g = \nabla_{\mathbf{U}}\Omega[Q(y|x), P(y|x)], \\
        \delta_{\mathbf{\Lambda}}^{*}\approx \varepsilon_2\frac{g}{||g||_{2}},\ \text{where}\ g = \nabla_{\mathbf{\Lambda}}\Omega[Q(y|x), P(y|x)].\
    \end{aligned}
\end{equation}
Note that $\varepsilon_1$ and $\varepsilon_2$ are hyperparameters controlling the magnitudes of perturbations on eigenvector and eigenvalue perturbations, respectively. Typically, $\varepsilon_1$ and $\varepsilon_2$ are small values to ensure the generated adversarial examples are ``unnoticeable''. Figure~\ref{fig:framework} presents the framework of AT for various types of perturbations, including traditional AT and our proposed SAT. SAT focuses on the graph structure input but allows continuous adversarial perturbations on the eigenpairs, incorporating the structural information of the graph and further improving the effectiveness of AT.

\subsection{Objective Function Definition}
For AT of GNNs, two adversarial regularization terms are added to the original loss to help improve the robustness against the adversarial perturbations. Then the final objective function is defined as follows:
\begin{equation}\label{eq:loss}
    \begin{aligned}
        \mathcal{L} = \mathcal{L}_\text{train} + \underbrace{ \alpha\mathcal{L}_{\text{SAT}}^{\text{vec}} + \beta\mathcal{L}_{\text{SAT}}^{\text{val}}}_{\text{adversarial regularizer}},
    \end{aligned}
\end{equation}
where $\alpha$ and $\beta$ are two trade-off hyperparameters adopted to balance the regularization terms. $\mathcal{L}_\text{train}$ is defined in Equation \eqref{eq:loss_train} where $\Omega=\text{CE}(Q(y,x), P(y,x)$ is the classification loss and $\Gamma=\gamma ||\Theta||^2$ is the regularization loss on $\Theta$. By jointly optimizing the classification loss and the regularization terms, we enforce the smoothness between the original graph and the adversarial graph, therefore improving the model's robustness.

Algorithm \ref{alg:training} summarizes the details of our proposed SAT method. Technically, the proposed approach is generalized to most GNNs by replacing the adjacency matrix with its rank-$r$ approximation based on spectral decomposition. The obtained eigenpairs that lie in the continuous space can be treated as low-dimensional embeddings of the graph. For each training loop, the perturbations on eigenvalues and eigenvectors are approximated based on Equation \eqref{eq:approx} to facilitate AT. Finally, the resulting model is expected to have high robustness against adversarial attacks while maintaining high classification accuracy.

\begin{algorithm}[t]
    \caption{Spectral Adversarial Training (SAT) for GNNs}
    \label{alg:training}
    \textbf{Input}: graph $G=(\mathbf{A}, \mathbf{X})$, adversarial regularization parameters $\alpha$ and $\beta$, norms of perturbations $\varepsilon _1$ and $\varepsilon _2$,  approximation rank $r$; \\
    \textbf{Output}: model parameter $\Theta$;
    \begin{algorithmic}[1] 
        \STATE initialize the model parameters $\Theta$;
        \STATE  eigenvalues $\mathbf{\Lambda}$ and eigenvectors $\mathbf{U}$ $\longleftarrow$ compute the rank-$r$ approximate of the normalized Laplacian matrix $\hat{\mathbf{A}}$;
        \STATE // criteria, such as early stopping for accuracy;
        \WHILE{stopping criteria is not met}
        \STATE // construct adversarial perturbations;
        \STATE
        $\delta_\mathbf{U}^{*}$, $\delta_{\mathbf{\Lambda}}^{*}\longleftarrow$ Equation (\ref{eq:approx});
        \STATE // compute adversarial regularization loss;
        \STATE $\mathcal{L}_{\text{SAT}}^{\text{vec}}$, $\mathcal{L}_{\text{SAT}}^{\text{val}} \longleftarrow$ Equation (\ref{eq:adv});
        \STATE update $\Theta$ by jointly optimizing classification loss and adversarial regularization loss in Equation (\ref{eq:loss}) with gradient descent method;
        \ENDWHILE
        \STATE \textbf{return} $\Theta$.
    \end{algorithmic}
\end{algorithm}

\subsection{Time and Memory Complexity}\label{sec:complexity}
\textbf{Time complexity.} Compared to standard training of GNNs, SAT additionally introduces two adversarial regularizers that promote the smoothness of GNNs over the given graph and make GNNs more robust against adversarial perturbations. The additional computation of SAT is twofold: (i) adversarial perturbations $\delta_\mathbf{U}^*$ and $\delta_\mathbf{\Lambda}^*$ calculated with Equation \eqref{eq:approx}, and (ii) adversarial regularizers $\mathcal{L}_{\text{SAT}}^{\text{vec}}$ and $\mathcal{L}_{\text{SAT}}^{\text{val}}$ calculated with Equation \eqref{eq:adv}. Firstly, calculating $\delta_\mathbf{U}^*$ and $\delta_\mathbf{\Lambda}^*$ requires one set of forward-propagation and back-propagation based on the linear approximation, where the forward-propagation has already been done and the result can be reused from the model. Secondly, SAT computes the two adversarial regularizers as additional smoothness constraints, respectively, which requires two forward propagations. Considering that we use the rank-$r$ approximation of the graph where $r\ll n$, the overall computation will be greatly reduced and the overhead of SAT is acceptable.

\textbf{Memory complexity.}
SAT requires additional memory to store the adversarial perturbations $\delta_\mathbf{U}^* \in \mathrm{R}^{n\times r}$ and $\delta_\mathbf{\Lambda}^*\in \mathrm{R}^{r\times r}$. However, the additional memory is insignificant since $r\ll n$. Furthermore, the original graph structure denoted by $\mathbf{A}\in \{0,1\}^{n\times n}$ is projected into a low-dimensional space using the rank-$r$ approximation, and it also reduces the time and memory required for optimization. Overall, the memory complexity of our algorithm is still dominated by the scale of the dataset (\emph{i.e.}, $n$). As a result, the memory usage of SAT is entirely acceptable in most cases.

\section{Experiments}\label{sec:exp}
\subsection{Experimental Settings}
\label{sec:setting}
\subsubsection{Datasets}
To evaluate the model performance and robustness on the semi-supervised node classification task, we conduct comprehensive experiments on four widely used datasets: Cora, Cora-ML, Citeseer, and Pubmed~\cite{mccallum2000automating,DBLP:journals/aim/SenNBGGE08}. We follow the previous work~\cite{zugner2018adversarial,li2021adversarial} and randomly split each dataset into training (10\%), validation (10\%), and testing (80\%) 
sets. The statistics of datasets are summarized in Table \ref{table:datastatistics}. Here only the largest connected component of each graph is considered, same as~\cite{zugner2018adversarial,li2021adversarial}.

\begin{table}[t]
    \centering
    \caption{Statistics of datasets used in our experiments. Note that only nodes and edges in the largest connected component of each graph are considered.}
    \begin{tabular}{lcccc}
        \toprule
        \textbf{}            & \textbf{Cora} & \textbf{Cora-ML} & \textbf{Citeseer} & \textbf{Pubmed} \\
        \midrule
        \midrule
        \textbf{\#Nodes}     & 2,485         & 2,810            & 2,100             & 19,717          \\
        \textbf{\#Edges}     & 5,069         & 7,981            & 3,668             & 44,324          \\
        \textbf{\#Features}  & 1,433         & 2,879            & 3,703             & 500             \\
        \textbf{\#Classes}   & 7             & 7                & 6                 & 3               \\
        \textbf{Avg. Degree} & 4.07          & 5.68             & 3.48              & 4.50            \\
        \bottomrule
    \end{tabular}
    \label{table:datastatistics}
\end{table}

\subsubsection{Baselines}
To comprehensively compare our proposed approach with several state-of-the-art methods, we consider the following four categories of methods: (i) Random and GCN-PGD from structured-based AT methods; (ii) BVAT, GraphVAT, and LAT from feature-based AT methods; (iii) Jaccard and SVD from adversarial purification methods and (iv) SimPGCN and MedianGCN from methods with robust architectures.
\begin{itemize}
    \item Random~\cite{DBLP:conf/icml/DaiLTHWZS18}: it is the simplest structure-based AT that randomly drops edges during each training step to generate adversarial examples.
    \item GCN-PGD~\cite{xu2019topology}: it is an attack-oriented AT that generates adversarial examples on the discrete graph structure based on projected gradient descent (PGD) attack.
    \item BVAT~\cite{deng2019batch}: it is a batch virtual AT that promotes the smoothness of GNNs by generating virtual adversarial perturbations.
    \item GraphVAT~\cite{feng2019graph}: it proposes a dynamic regularization technique based on virtual AT, to smooth the output and learn a robust GNN over adversarial examples.
    \item LAT~\cite{JinZ20}: it proposes to employ AT on the first latent representation of GNNs and make GNNs more robust against adversarial perturbations.
    \item Jaccard~\cite{wu2019adversarial}: since attackers tend to connect nodes with large differences to achieve a better attack effect, it uses Jaccard similarity to preprocess the graph and filter out edges between nodes with low feature similarity.
    \item SVD~\cite{entezari2020all}: it adopts a truncated SVD to form a low-rank approximation of graph adjacency matrix to alleviate the adversarial behaviors. SVD is similar to our method, but it does not incorporate adversarial training into the learning phase.
    \item SimPGCN~\cite{DBLP:conf/wsdm/JinDW0LT21}: it is an improved GNN with a feature similarity preserving aggregation, which effectively preserves node similarity while exploiting graph structure.
    \item MedianGCN~\cite{median}: it is a state-of-the-art defense method, adopting median as the aggregation function of GNNs to enhance their robustness.    
\end{itemize}

\subsubsection{SAT Setup}
For the proposed SAT approach, we have five important hyperparameters to tune: (i) $\varepsilon_1$ and $ \varepsilon_2$: the scales of adversarial perturbations on eigenvectors and eigenvalues, respectively; (ii) $\alpha$ and $\beta$: the weights for the two graph adversarial regularizers; (iii) $r$: the rank of the approximated graph. Particularly, we perform grid-search within a range of values and carefully tune the hyperparameters to balance classification power and model robustness. In this experiment, we set $\varepsilon_1 = \varepsilon_2=0.1$, $\alpha=\beta=0.5$ for all datasets, while $r=30$ for small-scale datasets Cora, Cora-ML, and Citeseer, and $r=150$ for large-scale dataset Pubmed (see detailed discussion in Section \ref{sec:exp3}). We employ SAT on three well-established GNNs, \emph{i.e.}, GCN, SGC and $\text{S}^2$GC detailed in Section \ref{sec:gnn}. The resulting models are named GCN-SAT, SGC-SAT, and $\text{S}^2$GC-SAT, respectively. For SGC and S$^2$GC, $K$ is set as 2 and 5 across all datasets, respectively. Additionally, we set $a=0.2$ for S$^2$GC. The number of training epochs is set to 100 and Adam optimizer~\cite{KingmaB14} is adopted to learn the model, with an initial learning rate of 0.01, 0.2, and 0.01 for GCN-SAT, SGC-SAT, and S$^2$GCN-SAT, respectively. 
It is worth noting that SAT is not limited to the above-mentioned three GNNs and that the technique can be straightforwardly generalized for other GNNs.

\subsubsection{Baseline Setup}
For Random, we set the edge dropping ratio as 0.5 during training; For GCN-PGD, we use PGD attack~\cite{xu2019topology} to generate adversarial examples, with perturbations rate set as 5\% of the edges and the optimization step set as 10 in each epoch. For BVAT, GraphVAT and LAT, we tune the scale of adversarial perturbations on the (hidden) features in the range \{0.01, 0.03, 0.05, 0.07, 0.1\} to achieve the best performance. For Jaccard and SVD, we empirically tune the threshold of dropping dissimilar edges from \{0, 0.01, 0.05, 0.08, 0.1\} and the approximation rank from 30 to 500. We finally set the threshold as 0.01 and the approximation rank the same as SAT across all datasets. All the baselines are trained with Adam optimizer with an initial learning rate of 0.01. We train them for 200 epochs with early stopping, where we stop training if there is no further improvement on the validation accuracy during 50 epochs. All these baselines use GCN as the backbone model according to the authors' implementations. For other hyperparameters, we empirically follow the default parameters in the original literature and fine-tune the model to achieve the best performance.

\subsubsection{Evaluation Protocol}
To evaluate the model robustness, we report the accuracy on the test set under the adversarial targeted attacks NETTACK~\cite{zugner2018adversarial} and PGD\footnote{Note that PGD is an untargeted attack method, we adapt the code provided by the authors to fit the targeted attack setting.}~\cite{xu2019topology}, for which 1,000 nodes are randomly selected as targeted nodes from the testing set. For PGD attack, the iteration step is set as 100 to achieve the best performance. Both methods take GCN as the surrogate model to conduct attacks.
We employ direct attacks where the perturbations are directly applied to connections of targeted nodes. We exactly follow the experimental settings in \cite{li2021adversarial}, which focus on the evasion attack scenario. Specifically, the perturbation budget is set as the node degrees. For all methods, we report classification accuracy averaged over 10 runs with different initial weights. We implement SAT with open-sourced software GraphGallery~\cite{li2021graphgallery}. Code to reproduce the experiments is publicly available at \url{https://github.com/EdisonLeeeee/SAT}.

\begin{table}[t]
\centering
    \caption{Classification accuracy (\%) of standard GNNs and their variants incorporated with SAT and SVD on the clean datasets.}
    \begin{tabular}{lcccc}
    \toprule
    \textbf{} & \textbf{Cora} & \textbf{Cora-ML} & \textbf{Citeseer} & \textbf{Pubmed} \\
    \midrule
    \midrule
    \textbf{GCN} & 86.6$_{\pm0.3}$ & 86.0$_{\pm0.2}$ & 71.2$_{\pm0.4}$ & 85.6$_{\pm0.2}$  \\
    \textbf{w / SAT} & 85.2$_{\pm0.2}$ & 86.1$_{\pm0.2}$ & 72.4$_{\pm0.4}$ & 85.8$_{\pm0.1}$  \\
    \textbf{w / SVD} & 78.0$_{\pm1.1}$ & 77.1$_{\pm0.8}$ & 68.6$_{\pm1.2}$ & 80.4$_{\pm0.6}$\\
    \midrule
    \textbf{SGC} & 84.7$_{\pm0.1}$ & 85.4$_{\pm0.2}$ & 72.5$_{\pm0.5}$ & 84.7$_{\pm0.4}$  \\
    \textbf{w / SAT} & 84.5$_{\pm0.3}$ & 85.8$_{\pm0.6}$ & 72.9$_{\pm0.3}$ & 85.4$_{\pm0.3}$  \\
    \textbf{w / SVD} & 76.9$_{\pm1.6}$ & 77.8$_{\pm0.8}$ & 68.1$_{\pm1.4}$ & 79.7$_{\pm0.9}$\\
    \midrule
    \textbf{S$^ 2$GC} & 87.0$_{\pm0.3}$ & 86.4$_{\pm0.5}$ & 73.6$_{\pm1.0}$ & 85.2$_{\pm0.1}$  \\
    \textbf{w / SAT} & 85.7$_{\pm0.5}$ & 85.5$_{\pm0.3}$ & 73.4$_{\pm0.6}$ & 85.4$_{\pm0.2}$  \\
    \textbf{w / SVD} & 78.7$_{\pm0.9}$ & 78.5$_{\pm0.7}$ & 68.9$_{\pm1.6}$ & 81.2$_{\pm0.9}$\\
    \bottomrule
    \end{tabular}
    \label{table:clean_acc}
\end{table}

\begin{figure*}[t]
    \centering
    \includegraphics[width=0.9\textwidth]{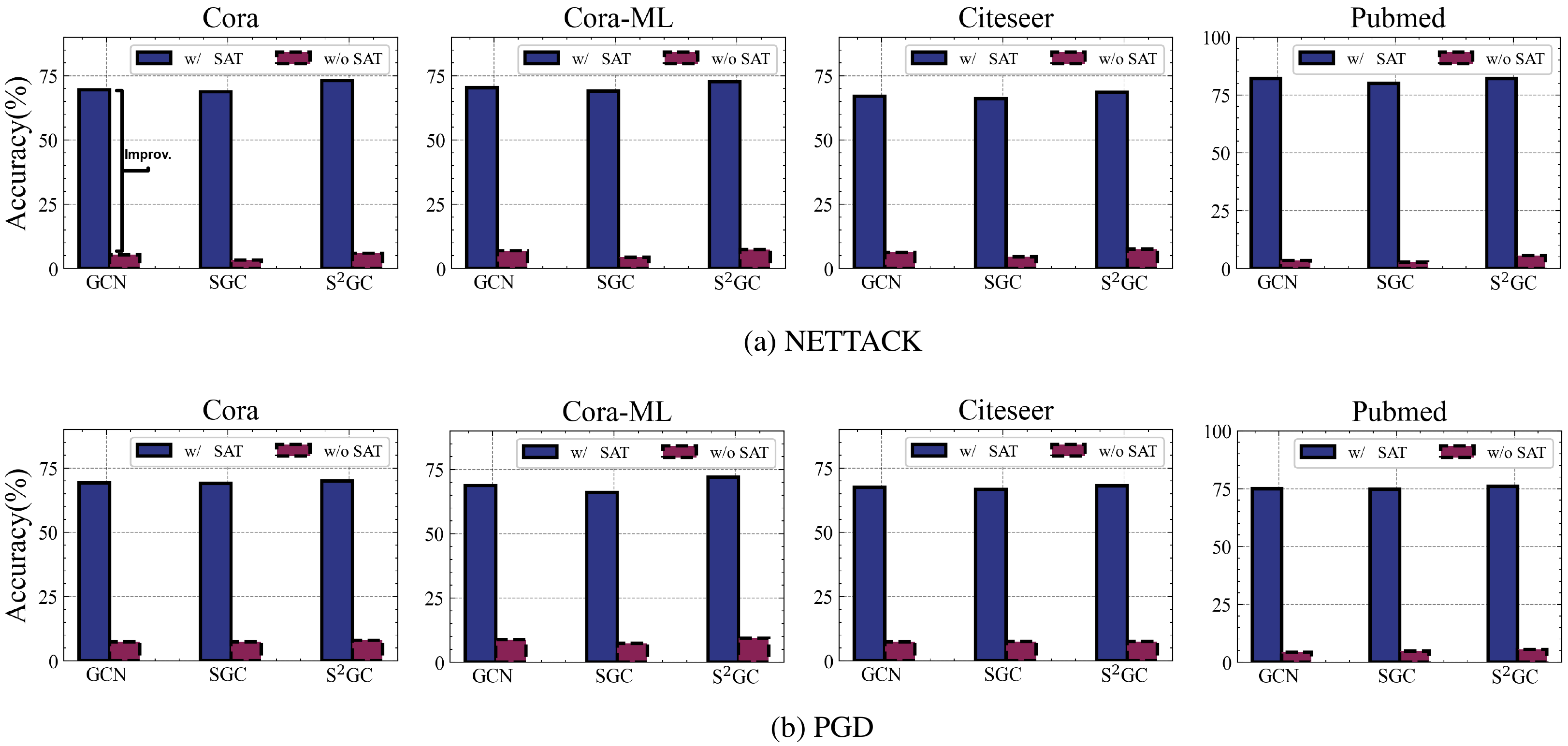}
    \caption{Robustness of standard GNNs w/ and w/o SAT on four datasets, measured by (a) NETTACK and (b) PGD attacks, respectively.}
    \label{fig:bar}
\end{figure*}

\subsection{Standard GNNs with SAT}\label{sec:exp1}
\subsubsection{Performance on Clean Data}
To empirically validate the effectiveness of our proposed SAT, we first investigate if the proposed algorithm will hinder the performance of standard GNNs on the clean datasets, \emph{i.e.}, the unperturbed datasets. In Table \ref{table:clean_acc}, we report the classification results of three GNNs, GCN, SGC, and S$^2$GC, on which standard training, SAT and SVD defense are employed, respectively. From the results, we have the following observations:
\begin{itemize}
    \item Although SAT adopts the rank-$r$ approximation of the graph structure and jointly optimizes the model with adversarial perturbations on the graph spectrum, the classification performance is not negatively affected in most cases. Note that SAT approximates the graph structure in a low-dimensional space, and thus leads to certain information loss. In general, the highest degradation is less than 2\%, which is acceptable in practice.
    \item Compared to SVD, a similar defense strategy that simply employs SVD as a low-rank preprocessing on the input graph, SAT achieves a better performance on the clean graphs. This suggests that although utilizing low-rank reconstruction would sacrifice the performance of the model on the graph, combining it with adversarial training can largely reduce such negative effects.
    \item SAT is also beneficial to promoting the generalization performance of GNNs. For instance, the results on SGC with SAT have demonstrated the effectiveness, in which the classification accuracy has a slight improvement in all datasets. It indicates that slight perturbations indeed help to improve the generalization ability, which is consistent with previous studies~\cite{feng2019graph,deng2019batch,hu2021robust}.    
\end{itemize}

\subsubsection{Robustness against Adversarial Attacks}
We further investigate the robustness of GNNs with SAT against adversarial attacks. We plot the results in Figure \ref{fig:bar}. The experimental results are obtained by conducting NETTACK and PGD, two strong adversarial structural attacks, on the targeted nodes with \emph{direct} attack setting. According to the results, we have the following observations:
\begin{itemize}
    \item Different GNNs show various robustness against adversarial attacks.  Specifically, SGC slightly underperforms GCN since it drops the nonlinearity between hidden layers, and a linear model is more vulnerable against attacks~\cite{goodfellow2014explaining}. We can also find that S$^2$GCN outperforms GCN and SGC across the four datasets. The main reason for this might be that NETTACK and PGD both take GCN as a surrogate model, leading to better attack performance on GCN than its linear variant SGC.
    \item A high level of robustness against the adversarial attack is achieved for GNNs with SAT. As can be seen, SAT remarkably improves the robustness of standard GNNs across all cases. We can find that the improvement of SAT is significant under both attacks, particularly PGD. Overall, the results affirmatively validate the claim that SAT does contribute to improving the robustness of GNNs. More specifically, the highlight of our approach is that SAT can greatly improve the robustness of GNNs while still maintaining classification accuracy.
\end{itemize}

\begin{table*}[t]
    \centering
    \caption{Overall robustness of various models when subjected to NETTACK and PGD attacks. The results are averaged over 10 runs with different initial weights. In each column, the \textbf{boldfaced} score denotes the best result.}
    \label{table:robustness}
    \resizebox{\textwidth}{!}{ 
        \begin{tabular}{cl*4{c}*4{c}}                                                                                                                                                                                                                                                                                                \\
            \toprule
            \textbf{}                           & \textbf{}          & \multicolumn{4}{c}{\textbf{NETTACK}} & \multicolumn{4}{c}{\textbf{PGD}}  \\
            \cmidrule{2-10}
            \textbf{}                           & \textbf{Method}     & \textbf{Cora}                              & \textbf{Cora-ML}                             & \textbf{Citeseer}      & \textbf{Pubmed}        & \textbf{Cora}           & \textbf{Cora-ML}        & \textbf{Citeseer}      & \textbf{Pubmed}         \\
            \midrule
            \midrule
            Standard GNNs                       & GCN & 5.3$_{\pm0.9}$ & 6.8$_{\pm0.8}$ & 6.3$_{\pm1.2}$ & 3.4$_{\pm0.3}$ & 7.0$_{\pm0.8}$ & 8.2$_{\pm0.9}$ & 7.4$_{\pm1.2}$ & 4.6$_{\pm0.5}$\\
            \midrule
            \multirow{2}{*}{Structure-based AT} & Random & 6.5$_{\pm1.7}$ & 7.3$_{\pm1.5}$ & 7.0$_{\pm1.9}$ & 4.2$_{\pm1.4}$ & 10.2$_{\pm1.6}$ & 9.5$_{\pm1.6}$ & 11.1$_{\pm1.7}$ & 10.0$_{\pm0.3}$\\
            \multicolumn{1}{c}{}                & GCN-PGD & 10.1$_{\pm1.6}$ & 9.8$_{\pm1.5}$ & 11.0$_{\pm1.5}$ & 8.6$_{\pm1.4}$ & 45.2$_{\pm1.7}$ & 50.4$_{\pm1.7}$ & 53.5$_{\pm1.9}$ & 62.3$_{\pm1.5}$\\
            \midrule
            \multirow{4}{*}{Feature-based AT}   & BVAT & 12.3$_{\pm1.4}$ & 11.5$_{\pm1.5}$ & 10.9$_{\pm1.5}$ & 9.6$_{\pm1.3}$ & 23.6$_{\pm1.5}$ & 19.4$_{\pm1.6}$ & 28.3$_{\pm1.6}$ & 15.7$_{\pm1.4}$\\
            \multicolumn{1}{c}{}                & GraphVAT & 10.4$_{\pm1.5}$ & 11.1$_{\pm1.6}$ & 10.7$_{\pm1.7}$ & 9.3$_{\pm1.3}$ & 20.7$_{\pm1.6}$ & 22.4$_{\pm1.6}$ & 30.3$_{\pm1.5}$ & 16.3$_{\pm1.3}$\\
            \multicolumn{1}{c}{}                & LAT &6.7$_{\pm1.7}$ & 7.4$_{\pm1.7}$ & 6.0$_{\pm1.9}$ & 4.4$_{\pm1.5}$ & 18.8$_{\pm1.8}$ & 17.0$_{\pm1.7}$ & 25.4$_{\pm1.8}$ & 14.7$_{\pm1.6}$\\
            \midrule
            \multirow{3}{*}{Other Defenses}      & Jaccard & 39.2$_{\pm1.2}$ & 31.1$_{\pm1.3}$ & 23.5$_{\pm1.3}$ & 54.2$_{\pm1.1}$ & 37.3$_{\pm1.2}$ & 28.8$_{\pm1.4}$ & 22.7$_{\pm1.4}$ & 40.8$_{\pm1.2}$\\
            \multicolumn{1}{c}{}                & SVD & {67.1$_{\pm1.4}$} & {68.3$_{\pm1.4}$} & {64.7$_{\pm1.5}$} & {77.5$_{\pm1.3}$} &{55.4$_{\pm1.4}$} & {58.7$_{\pm1.5}$} & {60.2$_{\pm1.4}$} & {70.9$_{\pm1.3}$}\\
            \multicolumn{1}{c}{}                & SimPGCN & 22.3$_{\pm1.2}$ & 34.5$_{\pm1.0}$ & 32.4$_{\pm1.3}$ & 45.6$_{\pm1.2}$ & 30.4$_{\pm1.5}$ & 39.3$_{\pm1.3}$ & 41.6$_{\pm1.2}$ & 47.9$_{\pm0.9}$\\            
            \multicolumn{1}{c}{}                & MedianGCN & 31.7$_{\pm1.5}$ & 43.3$_{\pm1.4}$ & 42.9$_{\pm1.5}$ & 59.8$_{\pm1.2}$ & 42.6$_{\pm1.3}$ & 50.2$_{\pm1.3}$ & 55.4$_{\pm1.6}$ & 59.2$_{\pm1.5}$\\
            \midrule
            \multirow{3}{*}{Ours}               & GCN-SAT & 69.5$_{\pm1.3}$ & 70.3$_{\pm1.5}$ & 67.1$_{\pm1.5}$ & 82.0$_{\pm1.4}$ & %
            69.6$_{\pm1.6}$ & 68.3$_{\pm1.5}$ & 67.0$_{\pm1.5}$ & 75.8$_{\pm1.5}$\\
            \multicolumn{1}{c}{}                & SGC-SAT & 68.7$_{\pm1.6}$ & 69.0$_{\pm1.5}$ & 66.9$_{\pm1.7}$ & 80.8$_{\pm1.3}$ %
            &69.0$_{\pm1.5}$ & 66.3$_{\pm1.7}$ & 66.7$_{\pm1.7}$ &74.7$_{\pm1.4}$\\
            \multicolumn{1}{c}{}                & $\text{S}^2$GC-SAT &\textbf{72.5$_{\pm1.5}$} & \textbf{71.6$_{\pm1.6}$}& \textbf{68.5$_{\pm1.6}$} & \textbf{82.6$_{\pm1.5}$} & %
            \textbf{70.2$_{\pm1.4}$} & \textbf{71.9$_{\pm1.7}$} & \textbf{65.1$_{\pm1.8}$} & \textbf{77.0$_{\pm1.5}$}\\
            \bottomrule
        \end{tabular}
    }
\end{table*}

\subsection{Robustness Comparison with Baselines}\label{sec:exp2}
In Section \ref{sec:exp1}, we empirically show the performance of different GNNs employed with SAT on four datasets. To further validate the robustness of our proposed approach, we adopt GCN-SAT, SGC-SAT, and S$^2$GC-SAT algorithms compared with multiple baselines for semi-supervised node classification on four datasets. The experimental results are detailed in Table \ref{table:robustness}. It should be noted that while the majority of the defense methods use GCN as the base model, we also report the result of GCN as a comparison baseline. In what follows, we summarized our major observations from the results.

GCN performs worst in all cases. The result is expected since the vulnerability of GNNs has been demonstrated in the literature. This points to an urgent need to improve the robustness of GNNs or design a more practical defense strategy. Random, as a simple structure-based AT method, slightly improves GCN's robustness by randomly dropping a subset of edges during training. The key to making this simple strategy work is that attackers prefer to add malicious edges rather than remove them, according to \cite{jinsurvey}. GCN-PGD is an attack-oriented AT method that outperforms Random in terms of robustness. For PGD attack, GCN-PGD shows better robustness since it generates adversarial perturbations based on the attacks that are essentially similar. However, it is less sufficient to resist a targeted attack, NETTACK. The experimental results highlight the issue that the attack-oriented method has poor generalization ability to resist an unseen attack.

Compared to structure-based AT, the feature-based AT methods, especially BVAT and GraphVAT, achieve better robustness against NETTACK. The results show that applying virtual AT to node features improves the robustness of GNNs. Surprisingly, LAT performs the worst in most cases, and its robustness even suffers a slight drop in Citeseer when compared to the original GCN. We believe that AT on the hidden representation is ineffective for defending against structural perturbations on inputs. Furthermore, the results support our claim that feature-based AT is insufficient to resist structural attacks because it ignores structural information and treats feature perturbations independently during training.

Beyond AT-based methods, different heuristics seek to improve the robustness of GNNs from various perspectives. Among these methods, MedianGCN and SimPGCN show fairly robust performance, demonstrating the efficacy of improving the robustness aggregation mechanism of GNNs. In addition,
Jaccard and SVD share similar motivation by preprocessing the input graph before training, which achieves better performance than most AT-based methods. More specifically, SVD is the most effective baseline method that outperforms other baselines in all cases. However, SVD is less effective when an adaptive attack, PGD, is performed. 

For our proposed SAT approach, the three models outperform other baselines in most cases. As can be seen, joint training with SAT can significantly improve GNNs' robustness by providing a simple way to resist adversarial attacks. As the only difference between GCN-SAT and GCN is applying the proposed SAT, the improvement is attributed to the proposed SAT, which greatly improves GCN's robustness. In particular, S$^2$GC-SAT boosts the performance most significantly and establishes state-of-the-art results across four datasets. 
The results verify that SAT is beneficial for GNNs' robustness.

Overall, we can conclude that the existing methods for improving GNNs' robustness through graph AT may be insufficient to resist powerful attacks like NETTACK. Our work suggests a way to improve the AT scheme from the spectral domain,  which is beneficial and meaningful for future research.

\begin{figure}[t]
    \centering
    \includegraphics[width=0.6\linewidth]{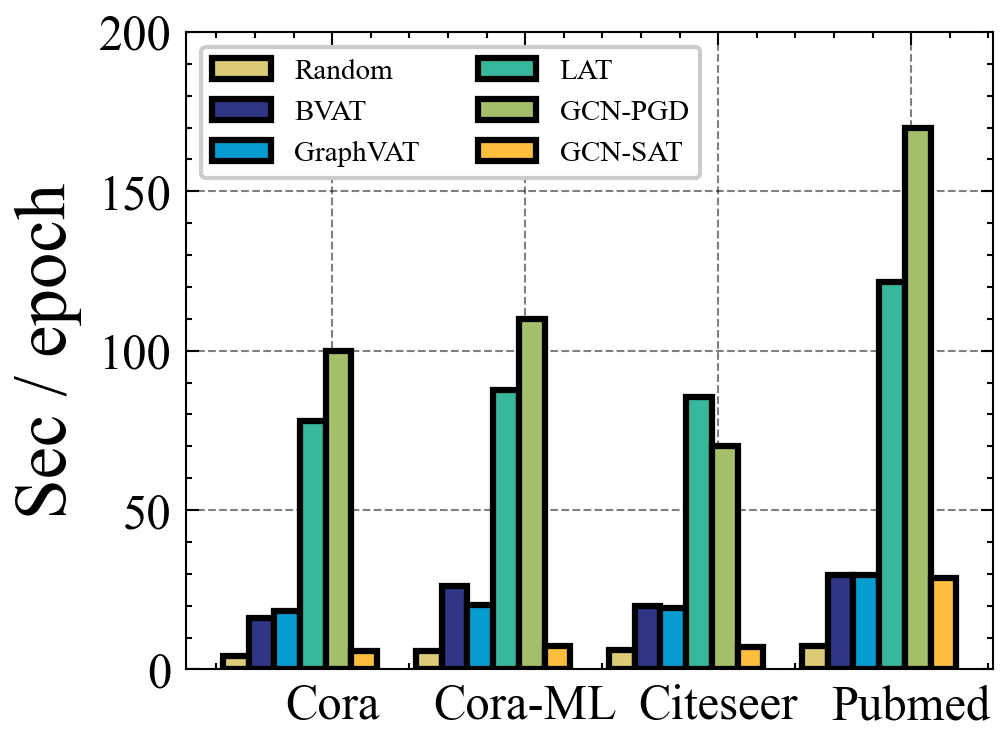}
    \caption{Running time comparison of GCN with various AT strategies in each training epoch. }
    \label{fig:time_bar}
\end{figure}

\subsection{Training Time Comparison with AT-based Methods}
In general, a good defense method has three characteristics: (i) less compromise on clean performance, (ii) higher robustness against adversarial attack, and (iii) acceptable training time. Clearly, SAT can help GNNs improve their robustness without sacrificing classification accuracy. In Section \ref{sec:complexity}, we briefly and theoretically analyze the time complexity of SAT. We now consider the practical running time of the algorithm. In Figure \ref{fig:time_bar}, we empirically estimate and compare the time consumption of GCN-SAT and other AT-based methods. The results are obtained on a machine with one NVIDIA RTX 2080Ti GPU and an Intel Core  i9-9900X CPU (3.50GHz). 

For these baselines, we have the following observations. (i) Random has little impact on the training efficiency of GCN since it is a random drop trick. (ii) BVAT and GraphVAT adopt a similar scheme, \emph{i.e.}, virtual AT, and therefore have similar performance in running time. (iii) LAT and GCN-PGD require a set of inner iterations to optimize the adversarial perturbations during each training step, resulting in dramatically high overheads compared with other methods.

According to Figure \ref{fig:time_bar}, the running time of GCN-SAT is slightly higher than Random, but much lower than other methods on Cora, Cora-ML, and Citeseer. As there are only two additional forward-propagations and one additional back-propagation, the training algorithm efficiency is not a problem in practical applications. In addition, the low-rank approximation benefits the scalability of SAT, which makes it capable of tackling the challenge of AT with large-scale datasets. In particular, GCN-SAT still shows acceptable running time in Pubmed. Generally, the results are consistent with our theoretical analysis, which demonstrates the efficiency of our proposed approach. Overall, the computation overhead for SAT is acceptable in most cases, and the training efficiency has not been negatively affected even when dealing with larger datasets such as Pubmed.

\begin{figure*}[th]
    \centering
    \includegraphics[width=\textwidth]{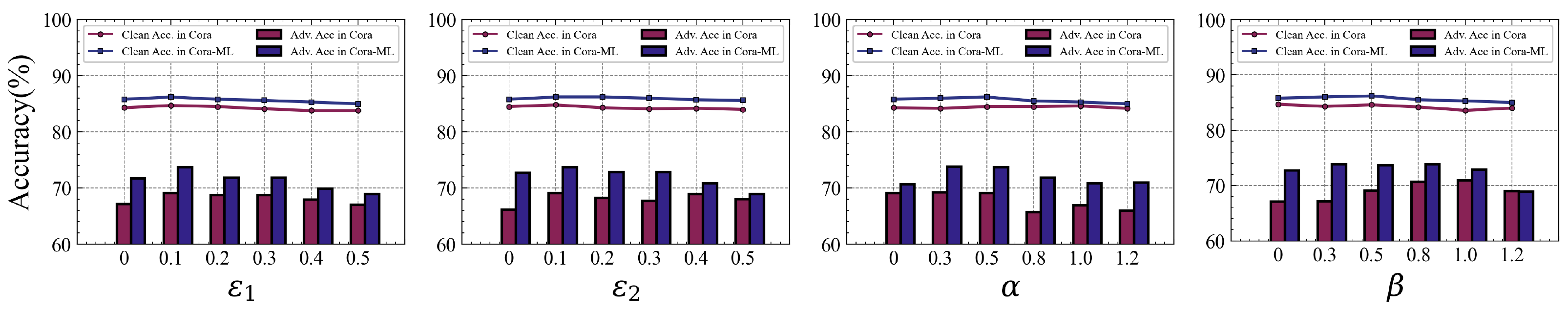}
    \caption{SAT components' ablation study \emph{w.r.t.} $\varepsilon_1$, $\varepsilon_2$, $\alpha$ and $\beta$. The performance of both clean and adversarial accuracy are presented. }
    \label{fig:abla}
\end{figure*}

\begin{figure*}[th]
    \centering
    \includegraphics[width=\textwidth]{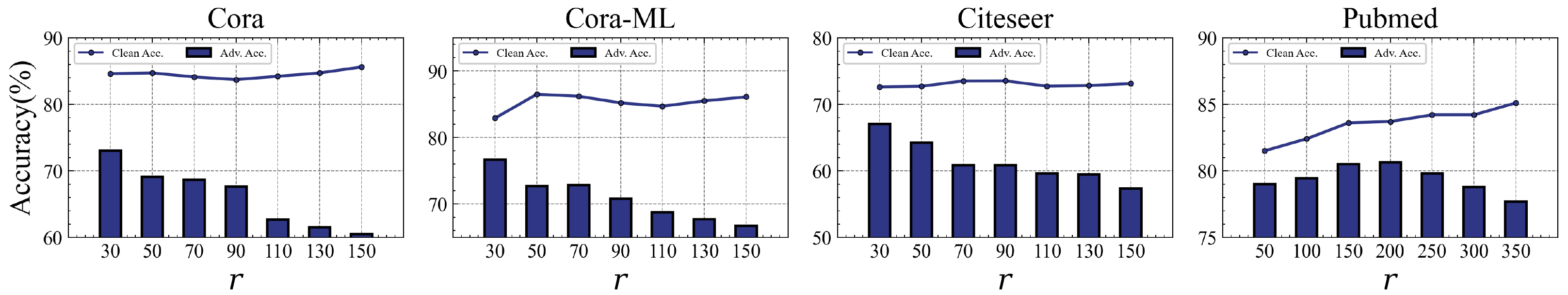}
    \caption{SAT components' ablation study \emph{w.r.t.} approximation rank $r$ on four datasets. The performance of both clean and adversarial accuracy are presented. }
    \label{fig:abla_r}
\end{figure*}

\subsection{Ablation Study}\label{sec:exp3}

\subsubsection{Parameter Sensitivity \emph{w.r.t.} $\varepsilon_1$, $\varepsilon_2$, $\alpha$ and $\beta$}
Recall that SAT has a set of hyperparameters: $\varepsilon_1$, $\varepsilon_2$, $\alpha$, $\beta$, and $r$, we next study the effects of different values of hyperparameters on the performance of the proposed SAT. We take GCN employed with SAT as our base model in the ablation study. 

We first investigate the effects of $\varepsilon_1$, $\varepsilon_2$, $\alpha$, and $\beta$, a GCN employed with SAT as our base model in the ablation study. Note that we only select GCN to tune the hyperparameters as SGC and S$^2$GC yield similar trends, and the results are omitted for brevity. Figure \ref{fig:abla} shows the model performance on Cora and Cora-ML, including the accuracy on the clean graph and the robustness against NETTACK on the perturbed graph. We use clean accuracy (clean Acc.) to refer to standard test accuracy (no adversarial perturbations) while adversarial accuracy denotes accuracy under attacks. For each hyperparameter, we vary the value in a range while the other hyperparameters are fixed with optimal values. 

Firstly, $\varepsilon_1$ and $\varepsilon_2$ control the magnitudes of perturbations on eigenvector and eigenvalue perturbations, respectively. In Figure \ref{fig:abla}, the clean accuracy and adversarial accuracy of GCN-SAT first increase and then decrease as $\varepsilon_1$ and $\varepsilon_2$ are increased from 0 to 0.5. It is reasonable since small perturbations may be beneficial for the model generalization ability, while larger perturbations would have the opposite effect. 
For the model robustness, enlarging $\varepsilon_1$ consistently leads to better robustness, but overlarge $\varepsilon_2$ would degrade the performance. 
Considering both clean accuracy and model robustness together, we set $\varepsilon_1=\varepsilon_2=0.1$ as optimal values.

Secondly, $\alpha$ and $\beta$ are two trade-off hyperparameters to balance the regularization terms. Observed from Figure \ref{fig:abla}, it is clear that both clean accuracy and model robustness achieve an increase in performance when $\alpha \leq  0.5$ and $\beta \leq 0.5$, which means that the two regularizers are important for training.  However, with the growth, the overall performance of the model would be negatively affected. Taking both clean accuracy and model robustness into account jointly, we set $\alpha=\beta=0.5$ as optimal values.

\subsubsection{Parameter Sensitivity \emph{w.r.t.} Approximation Rank $r$}
In addition to the above hyperparameters, $r$ is another important one to adjust for the model. The graph is approximated by the eigenpairs, where $r$ determines the richness of structural information. As demonstrated in \cite{entezari2020all}, imposing high-rank changes to the graph can be greatly alleviated by discarding the high-rank components of the graph. Following this insight, we adjust $r$ to achieve better classification performance as well as high robustness.

To demonstrate the importance of different $r$, we run an ablation study by varying $r$ in four datasets \emph{w.r.t.} NETTACK, and the results are displayed in Figure \ref{fig:abla_r}. As can be seen, with different values of $r$, GCN-SAT has a highly different sensitivity on the scale of the graph. In Cora, Cora-ML, and Citeseer, of which the graph is relatively small, we can see that a small number of $r$ is sufficient to capture the structural information and model the relationship between the nodes. Typically, a relatively good clean performance is achieved for $r=30$ and $r=50$ on the above three datasets. For a large-scale graph in Pubmed, a large $r$ is required to approximate the graph structure to avoid underfitting. The clean accuracy has a significant increase when $r$ is increased from 50 to 350.

As a coin has two sides, increasing $r$ has limitations as well as advantages. A large $r$ will lead to low robustness of the model since the perturbations are often high-rank~\cite{entezari2020all}. As shown in Figure \ref{fig:abla_r}, the robustness of the model is decreased, despite the classification accuracy being increased with larger $r$. The results in Cora demonstrate a clear trend especially when $r$ is greater than 90. For the Pubmed dataset, the robustness of the model could benefit from a relatively large $r$ as the classification performance has simultaneously improved. But when $r$ is greater than 250, the robustness is also slightly degraded due to being affected by the high-rank perturbations.

In addition to the model robustness, training efficiency is also important. A larger $r$ indicates more computational overhead is required in the two additional forward-propagations. Therefore, a trade-off needs to be made for our proposed approach. Thus, we choose $r=30$ for Cora, Cora-ML and Citeseer, while $r=150$ for Pubmed.

\subsubsection{Method Ablation}
\begin{table}[t]
    \centering
    \caption{Ablation studies of SAT on GNNs with different settings.}
    \resizebox{\linewidth}{!}{\begin{tabular}{lcccc}
        \toprule
        \textbf{}        & \textbf{Cora}  & \textbf{Cora-ML} & \textbf{Citeseer} & \textbf{Pubmed} \\
        \midrule
        \midrule
        & \multicolumn{4}{c}{\textbf{Clean}}\\
        \midrule        
        \textbf{GCN} & 86.6$_{\pm0.3}$ & 86.0$_{\pm0.2}$ & 71.2$_{\pm0.4}$ & 85.6$_{\pm0.2}$      \\
        \textbf{GCN-SAT}& 84.7$_{\pm0.2}$ & 86.2$_{\pm0.2}$ & 73.2$_{\pm0.4}$ & 85.5$_{\pm0.1}$\\
        \textbf{GCN-SAT ($\alpha=\beta=0$)} & 78.3$_{\pm1.5}$ & 77.4$_{\pm0.9}$ & 67.9$_{\pm1.5}$ & 79.8$_{\pm0.9}$ \\
        \textbf{SimPGCN-SAT}& 86.9$_{\pm0.7}$ & 87.1$_{\pm0.5}$ & 72.4$_{\pm0.6}$ & 86.4$_{\pm0.5}$\\           
        \midrule        
        & \multicolumn{4}{c}{\textbf{NETTACK}}\\        
        \midrule
        \textbf{GCN} & 5.3$_{\pm0.9}$ & 6.8$_{\pm0.8}$ & 6.3$_{\pm1.2}$ & 3.4$_{\pm0.3}$      \\
        \textbf{GCN-SAT}&69.5$_{\pm1.3}$ & 70.3$_{\pm1.5}$ & 67.1$_{\pm1.5}$ & 82.0$_{\pm1.4}$ \\
        \textbf{GCN-SAT ($\alpha=\beta=0$)} & 66.5$_{\pm1.3}$ & 67.8$_{\pm1.5}$ & 65.1$_{\pm1.5}$ & 77.0$_{\pm1.4}$ \\
        \textbf{SimPGCN-SAT}& 68.9$_{\pm1.1}$ & 70.1$_{\pm1.5}$ & 73.1$_{\pm1.5}$ & 79.9$_{\pm1.2}$\\          
        \midrule        
        & \multicolumn{4}{c}{\textbf{PGD}}\\
        \midrule
        \textbf{GCN} & 7.0$_{\pm0.8}$ & 8.2$_{\pm0.9}$ & 7.4$_{\pm1.2}$ & 4.6$_{\pm0.5}$      \\
        \textbf{GCN-SAT} & 69.6$_{\pm1.6}$ & 68.3$_{\pm1.5}$ & 67.0$_{\pm1.5}$ & 75.8$_{\pm1.5}$\\
        \textbf{GCN-SAT ($\alpha=\beta=0$)} & {56.1$_{\pm1.2}$} & {58.0$_{\pm1.0}$} & {61.4$_{\pm1.5}$} & {69.6$_{\pm1.7}$}\\
        \textbf{SimPGCN-SAT} & 71.6$_{\pm1.2}$ & 70.5$_{\pm1.0}$ & 69.3$_{\pm1.1}$ & 75.7$_{\pm1.7}$\\        
        \bottomrule
    \end{tabular}}
    \label{table:additional}
\end{table}

To evaluate the contribution of low-rank approximation and spectral adversarial training, we first conduct ablation studies on SAT built upon GCN with $\alpha=\beta=0$, \emph{i.e.}, performing SAT with only low-rank approximation. As shown in Table~\ref{table:additional}, the clean accuracy of SAT has decreased without adversarial training.  When adversarial attacks are performed, the robustness of GCN is still improved by SAT even without incorporating spectral adversarial training. In this regard, it achieves a similar performance with SVD. Indeed, the low-rank approximation contributes significantly to the robustness of GNNs, as suggested in \cite{entezari2020all}. When incorporating with the spectral adversarial training, the performance of GCN-SAT can be also boosted against attacks, especially the adaptive attack PGD. Overall, the results reveal that although the low-rank approximation is important for improving the adversarial robustness of GNNs, the adversarial training can improve the generalization performance of GNNs and is beneficial for defending against adaptive attacks such as PGD.

In addition, we can also observe from Table~\ref{table:additional} that SAT can also benefit SimPGCN, an improved GNN with robust architecture. 
The results demonstrate that combining different strategies with SAT further improves the model performance. Therefore, we believe that SAT can work as a general defense strategy when applied to different GNNs.

\section{Conclusion}\label{sec:conclusion}
We present the SAT method for improving GNN's resilience to adversarial perturbations, with a focus on the task of node classification. By studying the matrix perturbation theory, we approximate the low-rank graph structure by spectral decomposition and perform adversarial training on the eigenvalues and eigenvectors. In this way, we devise two regularizers to smooth the output distribution of the model and improve its performance. Extensive and rigorous experiments on four datasets demonstrate the effectiveness of our approach, showing that SAT can significantly improve the robustness of GNNs and outperform state-of-the-art competitors by a large margin. Furthermore, the SAT approach is generic to be employed on most GNN models without sacrificing classification power and training efficiency.

Studying the robustness of GNNs is an important problem, and this work provides essential insights for further research. In the future, we plan to study tasks beyond node classification and extend our research to other graph structures such as multiple and heterogeneous information graphs. Moreover, we would like to explore specific attacks such as untargeted attacks, poisoning attacks, and backdoor attacks, and investigate whether SAT is beneficial to improving the resistance of GNNs in such scenarios. These works, we believe, will be significant and influential in the future.

\ifCLASSOPTIONcompsoc
    \section*{Acknowledgments}
\else
    \section*{Acknowledgment}
\fi

The research is supported by the Key-Area Research and Development Program of Guangdong Province (2020B010165003), the Guangdong Basic and Applied Basic Research Foundation (2020A1515010831), the Guangzhou Basic and Applied Basic Research Foundation (202102020881), and CCF-AFSG Research Fund (20210002).

\ifCLASSOPTIONcaptionsoff
    \newpage
\fi

\bibliographystyle{IEEEtran}
\bibliography{main}

\begin{IEEEbiography}[{\includegraphics[width=1in,height=1.25in,clip,keepaspectratio]{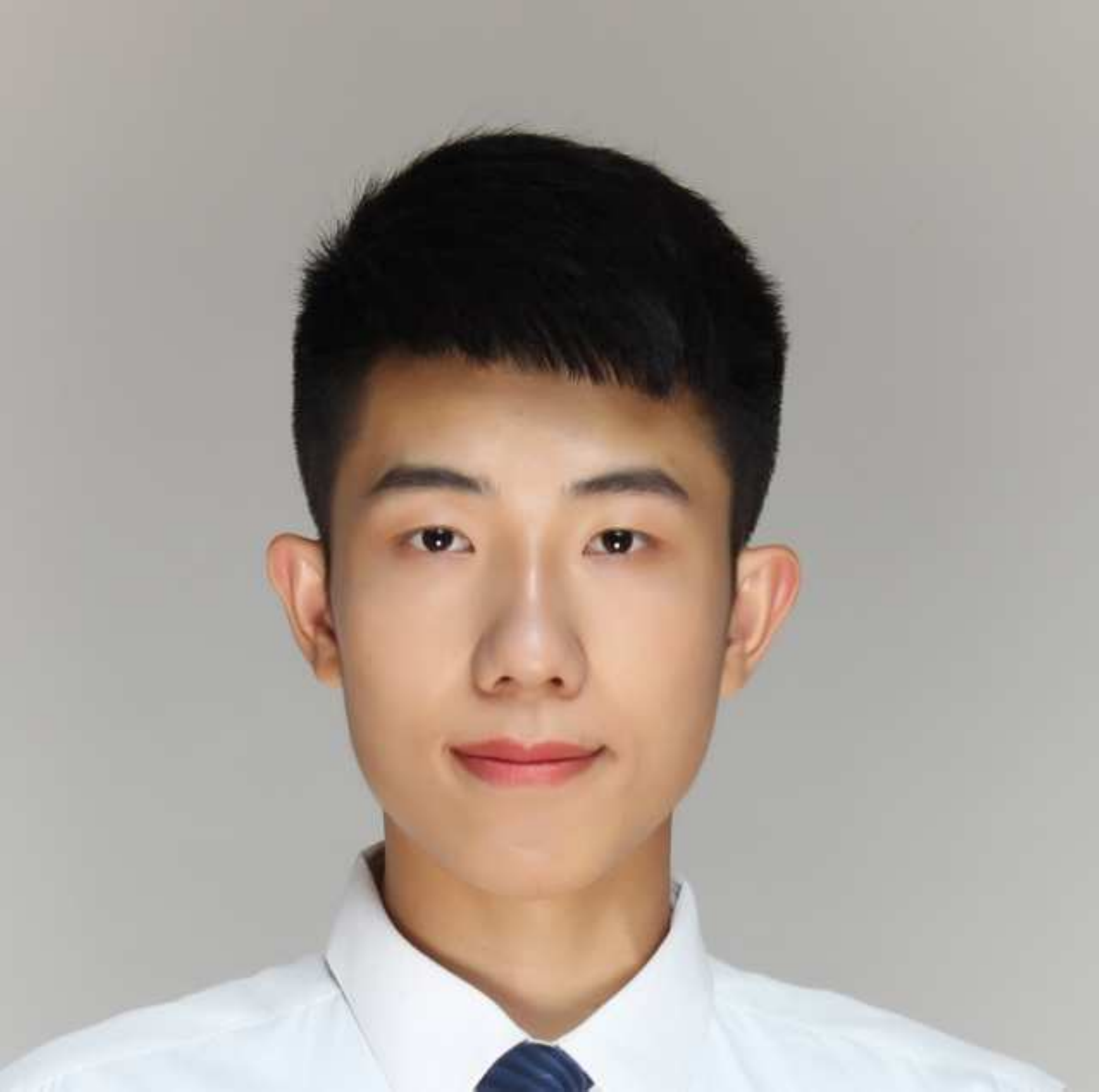}}]{Jintang Li}
  received the master's degree from Sun Yat-sen University (SYSU) in 2021. He is currently pursuing the Ph.D. degree with the School of Software Engineering, Sun Yat-sen University, Guangzhou, China. His main research interests include graph representation learning, adversarial machine learning, and data mining techniques.
\end{IEEEbiography}

\begin{IEEEbiography}[{\includegraphics[width=1in,height=1.25in,clip,keepaspectratio]{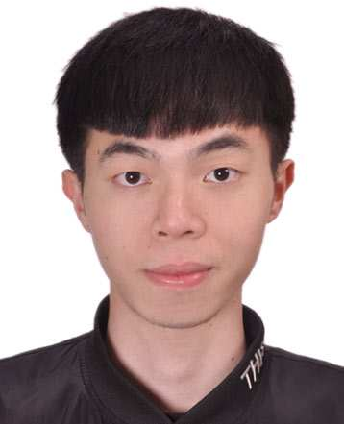}}]{Jiaying Peng}
   received the bachelor's and the master's degrees from South China Normal University and Sun Yat-sen University in 2019 and 2021, respectively. He is now working as a programmer and his interests include social networks, recommendation systems, machine learning, and data mining techniques.
\end{IEEEbiography}

\begin{IEEEbiography}[{\includegraphics[width=1in,height=1.25in,clip,keepaspectratio]{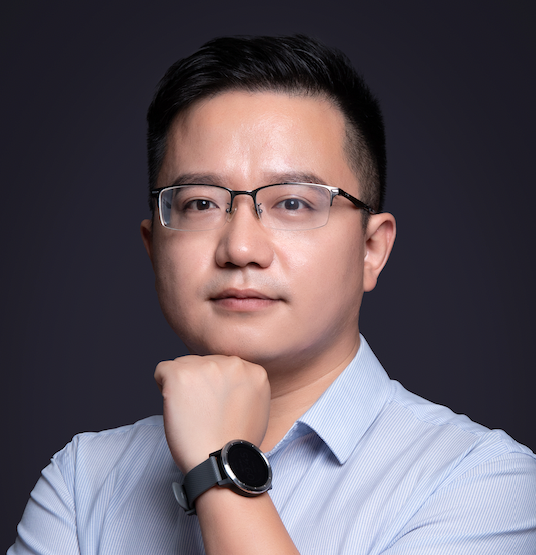}}]{Liang Chen}
  received the bachelor's and Ph.D. degrees from Zhejiang University (ZJU) in 2009 and 2015, respectively. He is currently an associate professor with the School of Computer Science and Engineering, Sun Yat-Sen University (SYSU), China. His research areas include data mining, graph neural network, adversarial learning, and services computing. In the recent five years, he has published over 70 papers in several top conferences/journals, including SIGIR, KDD, ICDE, WWW, ICML, IJCAI, ICSOC, WSDM, TKDE, TSC, TOIT, and TII. His work on service recommendation has received the Best Paper Award Nomination in ICSOC 2016. Moreover, he has served as PC member of several top conferences including SIGIR, WWW, IJCAI, WSDM etc., and the regular reviewer for journals including TKDE, TNNLS, TSC, etc.
\end{IEEEbiography}

\begin{IEEEbiography}[{\includegraphics[width=1in,height=1.25in,clip,keepaspectratio]{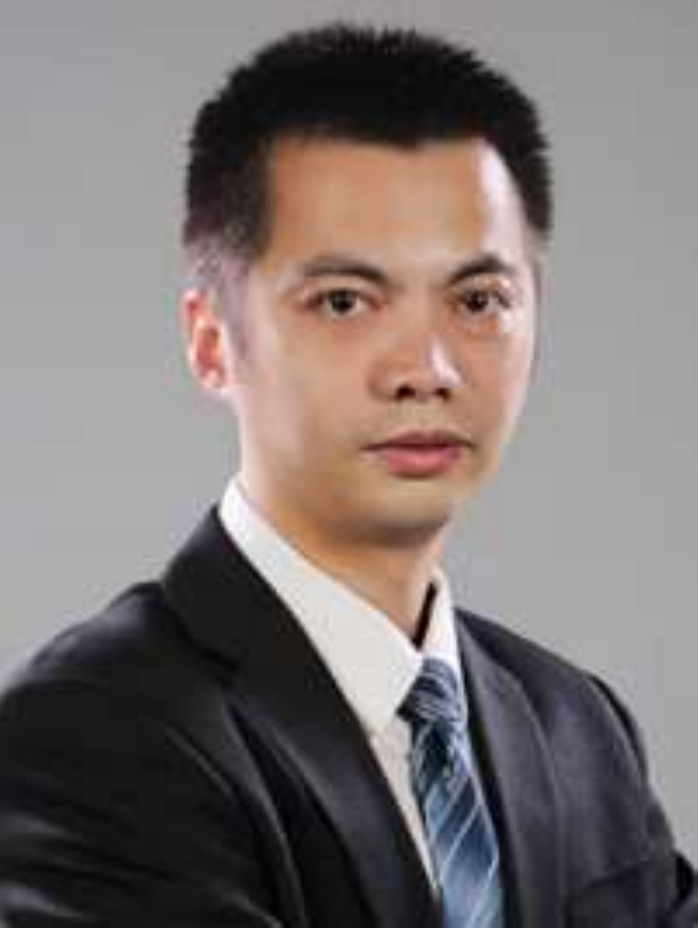}}]{Zibin Zheng}
  received the Ph.D. degree from the Chinese University of Hong Kong, in 2011. He is currently a Professor with the School of Computer Science and Engineering, Sun Yat-sen University, Guangzhou, China. His research interests include services computing, software engineering, and blockchain. He received the ACM SIGSOFT Distinguished Paper Award at the ICSE'10, the Best Student Paper Award at the ICWS'10, and the IBM Ph.D. Fellowship Award.
\end{IEEEbiography}

\begin{IEEEbiography}[{\includegraphics[width=1in,height=1.25in,clip,keepaspectratio]{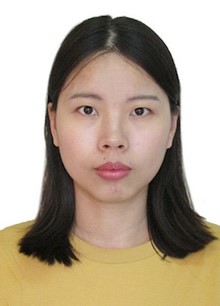}}]{Tingting Liang}
  received the Ph.D. degree from the College of Computer Science, Zhejiang University, in 2019. She is currently working as an Associate Professor in the School of Computer Science and Technology, Hangzhou Dianzi University. Her research interests include Data Mining, Recommender System, Multi-View Learning, Deep Learning, and Service Oriented Computing. Her papers have been published in some well-known conference proceedings and international journals such as ICDM, SIGIR, ICSOC, TITS, TSC, KBS, etc.
\end{IEEEbiography}

\begin{IEEEbiography}[{\includegraphics[width=1in,height=1.25in,clip,keepaspectratio]{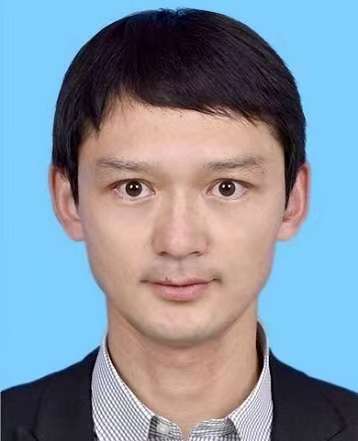}}]{Qing Ling}
  received the B.E. degree in automation and the Ph.D. degree in control theory and control engineering from the University of Science and Technology of China, Hefei, China, in 2001 and 2006, respectively. He was a Post-Doctoral Research Fellow with the Department of Electrical and Computer Engineering, Michigan Technological University, Houghton, MI, USA, from 2006 to 2009, and an Associate Professor with the Department of Automation, University of Science and Technology of China, from 2009 to 2017. He is currently a Professor with the School of Computer Science and Engineering, Sun Yat-sen University, Guangzhou, China. His current research interest includes distributed and decentralized optimization and its application in machine learning. His work received the 2017 IEEE Signal Processing Society Young Author Best Paper Award. Dr. Ling is a Senior Area Editor of IEEE SIGNAL PROCESSING LETTERS.

\end{IEEEbiography}
\end{document}